\documentclass[10pt,twocolumn,letterpaper]{article}

\PassOptionsToPackage{table}{xcolor}
\usepackage{cvpr}

\usepackage{microtype}
\usepackage{subcaption}
\usepackage{multirow}
\usepackage{colortbl}
\usepackage{tabularx}
\usepackage{pifont}
\usepackage{tikz}
\usepackage{placeins}
\usepackage{float}
\usetikzlibrary{positioning,arrows.meta,shapes.geometric,shadows,fit,calc}
\usepackage{mathtools}
\usepackage{amsthm}
\newtheoremstyle{tight}{0pt}{2pt}{\itshape}{}{\bfseries}{.}{.5em}{}
\theoremstyle{tight}
\newtheorem{proposition}{Proposition}
\usepackage{bbm}
\usepackage{algorithm}
\usepackage{algorithmic}

\newif\ifCoGAColoredTable
\AddToHook{env/table/begin}{\CoGAColoredTabletrue}
\AddToHook{env/table*/begin}{\CoGAColoredTabletrue}
\newcommand{\CoGATableStrut}{%
  \ifCoGAColoredTable
    \setbox\strutbox\hbox{\vrule
      height\dimexpr\ht\strutbox+0.1em\relax
      depth\dimexpr\dp\strutbox-0.1em\relax
      width0pt}%
  \fi
}
\AddToHook{env/tabular/begin}{\CoGATableStrut}
\AddToHook{env/tabular*/begin}{\CoGATableStrut}
\AddToHook{env/tabularx/begin}{\CoGATableStrut}

\newcommand{\ours}{CoGA}
\definecolor{headbg}{HTML}{F6D5E9}
\definecolor{setbg}{HTML}{DEEBF7}
\definecolor{oursbg}{HTML}{FCE4D6}
\definecolor{firstred}{HTML}{CA0000}
\definecolor{secgreen}{HTML}{1A9E77}
\definecolor{thirdblue}{HTML}{0066CC}
\newcommand{\fst}[1]{\textcolor{firstred}{\textbf{#1}}}
\newcommand{\snd}[1]{\textcolor{secgreen}{#1}}
\newcommand{\trd}[1]{\textcolor{thirdblue}{#1}}
\newcolumntype{Y}{>{\centering\arraybackslash}X}
\newcommand{\cmark}{\ding{51}}
\definecolor{deltapos}{HTML}{1A9E77}
\definecolor{deltaneg}{HTML}{CA0000}
\newcommand{\dpos}[1]{\textcolor{deltapos}{\textbf{#1}}}

\definecolor{cvprblue}{rgb}{0.21,0.49,0.74}
\usepackage[pagebackref,breaklinks,colorlinks,allcolors=cvprblue]{hyperref}

\def\confName{CVPR}
\def\confYear{2027}

\title{Intervene or Trust CLIP? Competence-Gated Adaptation for\\Source-Free Cross-Domain Few-Shot Learning}
\author{%
  {\normalfont\normalsize
  Haoyue Liu\textsuperscript{1,2}\quad
  Chenhao Zhu\textsuperscript{4}\quad
  Yu Huo\textsuperscript{1}\quad
  Haoran Shou\textsuperscript{1}\quad
  Shuguang Cui\textsuperscript{1,2,3}\quad
  Xiaoying Tang\textsuperscript{1,2,3,*}}\\[3pt]
  {\normalfont\small
  \textsuperscript{1}School of Science and Engineering, The Chinese University of Hong Kong, Shenzhen}\\
  {\normalfont\small
  \textsuperscript{2}Shenzhen Future Network of Intelligence Institute (FNii-Shenzhen)}\\
  {\normalfont\small
  \textsuperscript{3}Guangdong Provincial Key Laboratory of Future Networks of Intelligence, CUHK(SZ)}\\
  {\normalfont\small
  \textsuperscript{4}School of Computer Science, Beijing Jiaotong University, Weihai Campus}\\[1pt]
  {\normalfont\small
  \textsuperscript{*}Corresponding author.}\\[1pt]
  {\normalfont\small
  Emails: haoyueliu@link.cuhk.edu.cn, tangxiaoying@cuhk.edu.cn}
}
\begin{document}
\maketitle

\begin{abstract}
Source-free cross-domain few-shot learning (SF-CDFSL) adapts a pre-trained vision-language model such as CLIP to a distant target domain from a handful of labeled images, without access to source data. Recent methods improve the fine-tuning of CLIP with token-level interventions and apply them in the same way to every target domain. In this paper, we find that whether an intervention helps depends on how well frozen CLIP already recognizes the target domain: in 5-way 1-shot tasks, intervening improves accuracy by 0.72 points where frozen CLIP misclassifies most support images, but lowers it by 1.31 points where frozen CLIP already classifies them well, and the support set of a single episode suffices to tell the two cases apart; a bias-variance analysis explains both. Based on this finding, we propose Competence-Gated Adaptation (\ours{}), which measures frozen CLIP's accuracy and margin on the labeled support set and decides, before training, whether to intervene on the visual tokens or to trust CLIP by fusing the adapted prediction with the frozen one. \ours{} adds no trainable parameters and uses no query labels. On the four BSCD-FSL target domains, \ours{} achieves the best average accuracy, 59.56\% (1-shot) and 68.96\% (5-shot), outperforms ATHA and TIR by at least 2.82 and 1.39 points on the same episodes, and consistently improves three further backbones, three parameter-efficient fine-tuning frameworks and five further target domains.
\end{abstract}

\section{Introduction}
\label{sec:intro}

\begin{figure}[t]
  \centering
  \includegraphics[width=\columnwidth]{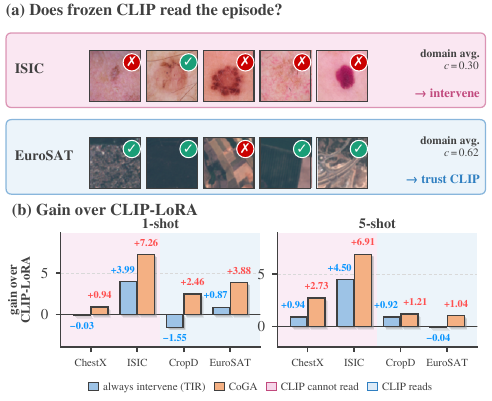}
  \vspace{-6mm}
  \caption{When to intervene on CLIP. (a) Frozen CLIP's predictions on two real 1-shot episodes and its average support accuracy $c$ per domain: it cannot read ISIC but reads EuroSAT. (b) Mean accuracy gains over CLIP-LoRA, computed from \cref{tab:main}. \ours{} improves all eight domain--shot settings; TIR lowers 1-shot CropDiseases by 1.55 points.}
  \label{fig:teaser}
  \vspace{-3mm}
\end{figure}

Cross-domain few-shot learning (CDFSL) recognizes new classes in a distant target domain, such as dermoscopy, satellite or chest X-ray images, from one or a few labeled examples per class~\citep{guo2020broader,tseng2020cross}. Its source-free variant (SF-CDFSL) adapts the model on the target support set alone~\citep{yazdanpanah2022visual,xu2024enhancing}. Here, fine-tuning CLIP~\citep{radford2021learning} with low-rank adapters (CLIP-LoRA~\cite{zanella2024low}) already rivals methods trained on source data~\citep{yi2026improving}.

Recent SF-CDFSL methods improve CLIP-LoRA by intervening on the visual tokens during fine-tuning: ATHA~\citep{yi2026improving} pushes the tokens least related to every class text away from the text features, and TIR~\citep{yi2026addressing} lowers the weight of tokens that all classes find similar and raises that of discriminative ones. Both apply the same intervention to every target domain. Yet target domains differ in how much CLIP knows about them: in 5-way 1-shot episodes, frozen CLIP classifies 62\% of the support images of EuroSAT and CropDiseases correctly, but only 30\% on ISIC and 21\% on ChestX, close to chance (\cref{fig:teaser}a). An intervention meant for domains CLIP cannot read may be unnecessary, or even harmful, on those it reads.

We test this directly in \cref{sec:analysis}. On the same episodes, we adapt CLIP once with a token intervention and once without it while fusing with frozen CLIP. In 1-shot episodes, intervening gains 0.72 points where frozen CLIP's support accuracy is below 0.5 and loses 1.31 points where it is above (\cref{fig:flip}), and across nine target domains, its gain falls as frozen CLIP's competence rises (\cref{fig:dom}). A bias-variance analysis explains why: trusting frozen CLIP pays when its bias on the domain is small against the noise of a few shots, and its competence on the support set indicates this bias (\cref{sec:theory}). This leads to the question we study:

\begin{center}
\emph{Can we decide, from the labeled support set alone,\\ when to intervene on CLIP and when to trust it?}
\end{center}

To answer it, we propose Competence-Gated Adaptation (\ours{}). Before training on an episode, \ours{} runs frozen CLIP once on the support images and measures its accuracy and margin. When both indicate that CLIP cannot read the episode, the gate opens, and the adaptation applies our updated token reweighting with centred anchors and a larger learning rate; otherwise, the adaptation runs without intervention and its prediction is fused with frozen CLIP's. The gate's threshold of 0.5 is fixed for all domains and shots. Description anchors, online crops and a trajectory ensemble complete \ours{}. As shown in \cref{fig:teaser}b, \ours{} improves on CLIP-LoRA in all eight domain--shot settings, by up to 7.26 points; on 1-shot CropDiseases, TIR lowers accuracy by 1.55 points while \ours{} raises it by 2.46 points. We summarize the contributions as follows:
\begin{itemize}\setlength{\itemsep}{1pt}\setlength{\topsep}{2pt}
  \item We first show that whether a token intervention helps CLIP in SF-CDFSL depends on the target domain: it helps where frozen CLIP cannot read the domain and hurts where it can, and a single labeled support set already tells the two cases apart. A simplified bias-variance analysis explains this dependence.
  \item We propose \ours{}, which replaces the uniform intervention of prior methods with a competence gate: before training, the gate chooses from the support set alone between intervening on the visual tokens and trusting frozen CLIP by fusing its prediction. The gate adds no trainable parameters, uses no query labels, and can be integrated with other token-level intervention methods, as demonstrated with ATHA.
  \item On BSCD-FSL, \ours{} achieves the best average accuracy, 59.56\% (1-shot) and 68.96\% (5-shot), outperforming all compared baselines by at least 2.82 and 0.90 percentage points, respectively. Without retuning, it also improves three further backbones, three fine-tuning frameworks, 7- and 10-way tasks and five further target domains.
\end{itemize}

\section{Related Work}
\label{sec:related}

\textbf{Cross-domain few-shot learning.} BSCD-FSL~\citep{guo2020broader} evaluates transfer from natural images to four distant domains. Methods that train on source data use style augmentation~\citep{fu2023styleadv}, flatter loss landscapes~\citep{zou2024flatten}, masked image modeling~\citep{ma2025reconstruction} or revised vision-transformer components~\citep{zou2024closer,zou2024attention,yi2025revisitingcontinuityimagetokens,yi2025random}. Source-free methods adapt on the target support set only~\citep{yazdanpanah2022visual,xu2024enhancing,xu2025step}. With CLIP as the backbone, recent work intervenes on the visual tokens~\citep{yi2026improving,yi2026addressing}, reuses intermediate text layers~\citep{zhang2026reclaiming}, keeps visual features discriminative~\citep{zhang2026mind} or rectifies local alignment~\citep{zhao2026interpretable}. These methods apply their intervention to every target domain. \ours{} decides whether to intervene with our updated reweighting; the gate also works with ATHA's alignment (\cref{sec:q3}).

\textbf{Few-shot adaptation of CLIP.} CLIP can be adapted through prompt learning~\citep{zhou2022learning,zhou2022conditional,khattak2023maple}, cache-based adapters~\citep{zhang2022tip,tang2024amu}, linear probes~\citep{huang2024lp++}, logit correction~\citep{li2025logits} and low-rank fine-tuning~\citep{zanella2024low}. Our focus is complementary to these prior adaptation mechanisms: \ours{} decides from the labeled support set whether visual-token intervention is warranted or frozen CLIP's prediction should be retained.

\textbf{Class descriptions and ensembles.} Class descriptions of visual properties improve zero-shot CLIP~\citep{menon2022visual,pratt2023does}, and snapshot ensembles~\citep{huang2017snapshot} average predictions along one training run. \ours{} uses both at no extra training cost.

\section{When Does Intervention Help?}
\label{sec:analysis}

\subsection{Preliminaries}

\textbf{SF-CDFSL.} A target task is an $N$-way $K$-shot episode with a support set $\mathcal{S}=\{(x_j,y_j)\}_{j=1}^{NK}$ and a query set $\mathcal{Q}$ drawn from the same $N$ classes. The model is adapted on $\mathcal{S}$ only and evaluated on $\mathcal{Q}$; the released loader of~\citet{yi2026improving,yi2026addressing} turns each support image into five views, and we write $\mathcal{S}_{\mathrm{aug}}$ for these $5NK$ labeled views. Following~\citet{yi2026improving,yi2026addressing}, we use CLIP with image encoder $f$ and text encoder $g$, and represent class $k$ by a unit-norm text anchor $t_k$ computed from prompts that contain the class name (\cref{sec:method}); $\mathrm{norm}(u)=u/\|u\|$. The prediction is $p(k\mid x)\propto\exp\big(s\cos(f(x),t_k)\big)$ with logit scale $s$. CLIP-LoRA~\citep{zanella2024low} adds low-rank adapters to the attention projections of both encoders and trains them with cross-entropy on $\mathcal{S}$.

\textbf{Token interventions.} We use our updated token reweighting in the paired analysis. Let $z_i$, $i=1,\dots,L$, be the patch tokens after a block of the image encoder ([CLS] not counted in $L$) and $u_i$ their projection into the joint embedding space. For each class $k$, our rule selects the set $\mathcal{T}_k$ of the $\lfloor\rho L\rfloor$ tokens most similar to the centred anchor $\tilde t_k$ (\cref{sec:open}), counts how many classes select each token, $S_i=\sum_{k}\mathbbm{1}[i\in\mathcal{T}_k]$, and rescales the token as $z_i\leftarrow w_i z_i$ with
\begin{equation}
  w_i=\begin{cases}1-\beta\,(S_i-s_0), & S_i>0,\\ 1, & S_i=0,\end{cases}
  \label{eq:rule}
\end{equation}
where $\rho$ is the fraction of tokens each class selects, $\beta$ the reweighting strength and $s_0$ the count that leaves a token unchanged ($s_0=3$ for 5-way tasks). Tokens selected by most classes are suppressed; tokens selected by a single class are amplified. Centring makes this selection depend on what distinguishes the class anchors. ATHA~\citep{yi2026improving} instead pushes the least similar tokens away from the text features. We call such a rule an \emph{intervention}.

\begin{figure*}[t]
  \centering
  \begin{subfigure}[b]{0.325\textwidth}
    \centering
    \includegraphics[width=\linewidth]{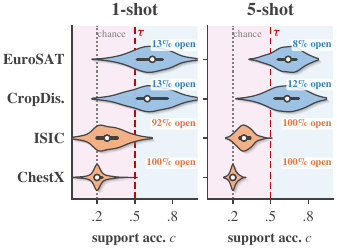}
    \caption{Competence $c$ per domain}
    \label{fig:comp}
  \end{subfigure}\hfill
  \begin{subfigure}[b]{0.325\textwidth}
    \centering
    \includegraphics[width=\linewidth]{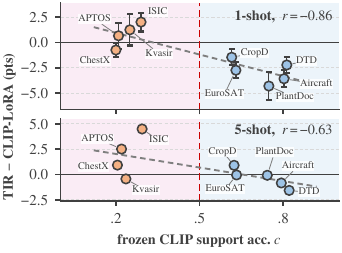}
    \caption{TIR's gain over CLIP-LoRA, nine domains}
    \label{fig:dom}
  \end{subfigure}\hfill
  \begin{subfigure}[b]{0.325\textwidth}
    \centering
    \includegraphics[width=\linewidth]{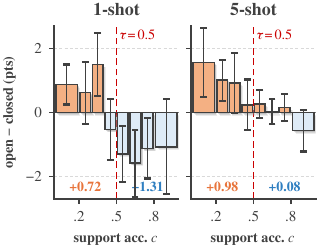}
    \caption{Open minus closed run, per episode}
    \label{fig:flip}
  \end{subfigure}
  \vspace{-2mm}
  \caption{Intervening improves accuracy at low frozen CLIP support accuracy in both shot settings. At high support accuracy, it reduces accuracy in 1-shot and has little effect in 5-shot. (a) Frozen CLIP's support accuracy $c$ separates the BSCD-FSL domains (all evaluation episodes; bar: interquartile range; dot: median; dashed: $\tau=0.5$; labels: gate open rate). (b) TIR gain versus average $c$ (diagnostic sets in \cref{fig:domapp} of the supplement). (c) Open minus closed run, binned by $c$ (200 episodes per BSCD-FSL domain); bottom: averages on each side of $\tau=0.5$. Error bars in (b,c): paired 95\% intervals where available; the 5-shot panel in (b) shows means only.}
  \label{fig:analysis}
  \vspace{-2mm}
\end{figure*}

\subsection{Frozen CLIP's Competence}

For an episode, we measure how well frozen CLIP reads it by its accuracy on the support images,
\begin{equation}
  c=\frac{1}{|\mathcal{S}_{\mathrm{aug}}|}\sum_{(x,y)\in\mathcal{S}_{\mathrm{aug}}}\mathbbm{1}\Big[\arg\max_{k}\cos\big(f_0(x),t^0_k\big)=y\Big],
  \label{eq:comp}
\end{equation}
and by its mean margin between the true class and the strongest wrong class,
\begin{equation}
  \begin{split}
  m=\frac{1}{|\mathcal{S}_{\mathrm{aug}}|}\sum_{(x,y)\in\mathcal{S}_{\mathrm{aug}}}\Big[&\cos\big(f_0(x),t^0_y\big)\\
  &-\max_{k\neq y}\cos\big(f_0(x),t^0_k\big)\Big],
  \end{split}
  \label{eq:margin}
\end{equation}
where $f_0$ is the frozen image encoder and $t^0_k$ the anchor computed with the frozen text encoder $g_0$. Both use only support labels. On ChestX and ISIC, frozen CLIP's similarities to the true-class and competing anchors are hard to distinguish; on CropDiseases and EuroSAT they separate distinctly (\cref{fig:sim} in the supplement).

Competence differs across domains (\cref{fig:comp}): at 1 shot, ISIC and ChestX episodes lie almost entirely below $c=0.5$, whereas EuroSAT and CropDiseases lie mostly above it. The gate can therefore decide from each episode's support set without a supplied domain label.

\subsection{Observation}

\textbf{Setup.} On the first 200 evaluation episodes of each BSCD-FSL domain, we adapt CLIP twice with identical seeds. The ``open'' run intervenes: it applies \cref{eq:rule} at blocks 8 and 10 and doubles the learning rate of the adapters. The ``closed'' run trusts frozen CLIP: it adapts without intervention and averages its prediction with frozen CLIP's. Everything else follows \ours{} (\cref{sec:setup}).

\textbf{Result.} As shown in \cref{fig:flip}, the effect of intervening changes sign with frozen CLIP's competence. In 1-shot episodes, below $\tau=0.5$, intervening gains $0.72\pm0.42$ points on average over 479 episodes; above, it loses $1.31\pm0.51$ over 321 episodes. With five shots, intervening still gains $+0.98\pm0.43$ below 0.5 and becomes harmless above it ($+0.08\pm0.23$): the cost of an unnecessary intervention falls as labels grow. The matched 1-shot comparison of TIR against CLIP-LoRA gives $+0.78\pm0.74$ points where $c<0.5$ and $-1.92\pm0.84$ where $c\geq0.5$. Across nine target domains (\cref{fig:dom}), TIR gain is negatively correlated with frozen CLIP's competence ($r=-0.86$ at 1 shot and $-0.63$ at 5 shots). In the 1-shot paired diagnostic, the gain is negative on all five domains with $c>0.5$.

\subsection{Why Competence Decides}
\label{sec:theory}

We explain the observation with a model of an $N$-way $K$-shot episode that compares the two branches by how much they trust frozen CLIP and measures the error of the class weights. The $d$-dimensional features of class $k$ are $f(x)=\mu_k+\xi$ with $\xi\sim\mathcal{N}(0,\sigma^2 I_d)$, and frozen CLIP's anchor is $t^0_k=\mu_k+b_k$ with a bias $b_k$ of norm $\delta$. Adaptation estimates $\mu_k$ by the mean $\hat\mu_k$ of $K$ support features, with error $v=d\sigma^2/K$, and trusting frozen CLIP to a degree $\gamma\in[0,1]$ gives the class weight $w_k(\gamma)=(1-\gamma)\hat\mu_k+\gamma\,t^0_k$. The closed branch, which keeps frozen CLIP's prediction, corresponds to a larger $\gamma$ than the open branch: $\gamma_c>\gamma_o$.

\begin{proposition}
\label{prop:shrink}
The error of the class weight is $R(\gamma)=\mathbb{E}\|w_k(\gamma)-\mu_k\|^2=(1-\gamma)^2v+\gamma^2\delta^2$, and intervening has the lower error if and only if
\begin{equation}
  \delta^2>\kappa\,\frac{d\sigma^2}{K},\qquad \kappa=\frac{2-\gamma_o-\gamma_c}{\gamma_o+\gamma_c}.
  \label{eq:cond}
\end{equation}
\end{proposition}

Trusting CLIP thus pays when its bias is small against the noise of $K$ shots. The bias is not observed, but frozen CLIP's competence indicates it: for a given class separation and noise, the expected nearest-anchor accuracy and squared-distance margin of frozen CLIP on the support set both decrease strictly in $\delta$, so the open branch has the lower error exactly when $\mathbb{E}[c]$ falls below a threshold $c^\star_K$ that grows with $K$ (\cref{prop:comp} in the supplement). Intervening accordingly pays on more episodes as shots grow (\cref{fig:flip,tab:switch}), and competence separates the domains where intervening helps from those where it hurts (\cref{fig:analysis}). As $c^\star_K$ depends on unobserved parameters, the gate applies a fixed $\tau$ to the observed $c$ together with the margin condition, a design that \cref{sec:q3} validates.

\textbf{Takeaways.} Intervening helps on the domains frozen CLIP cannot read and hurts on those it reads; which case holds is visible before training, in frozen CLIP's accuracy and margin on one support set; and where frozen CLIP reads the domain, its predictions are worth keeping. These motivate the gate (\cref{sec:gate}), the open branch (\cref{sec:open}) and the closed branch (\cref{sec:closed}) of \ours{}.

\section{Competence-Gated Adaptation}
\label{sec:method}

\ours{} decides from the support set whether to intervene and adapts with the matching branch (\cref{fig:overview}; \cref{alg:coga} in the supplement).

\begin{figure*}[t]
\centering
\includegraphics[width=\textwidth,height=17311468sp,keepaspectratio]{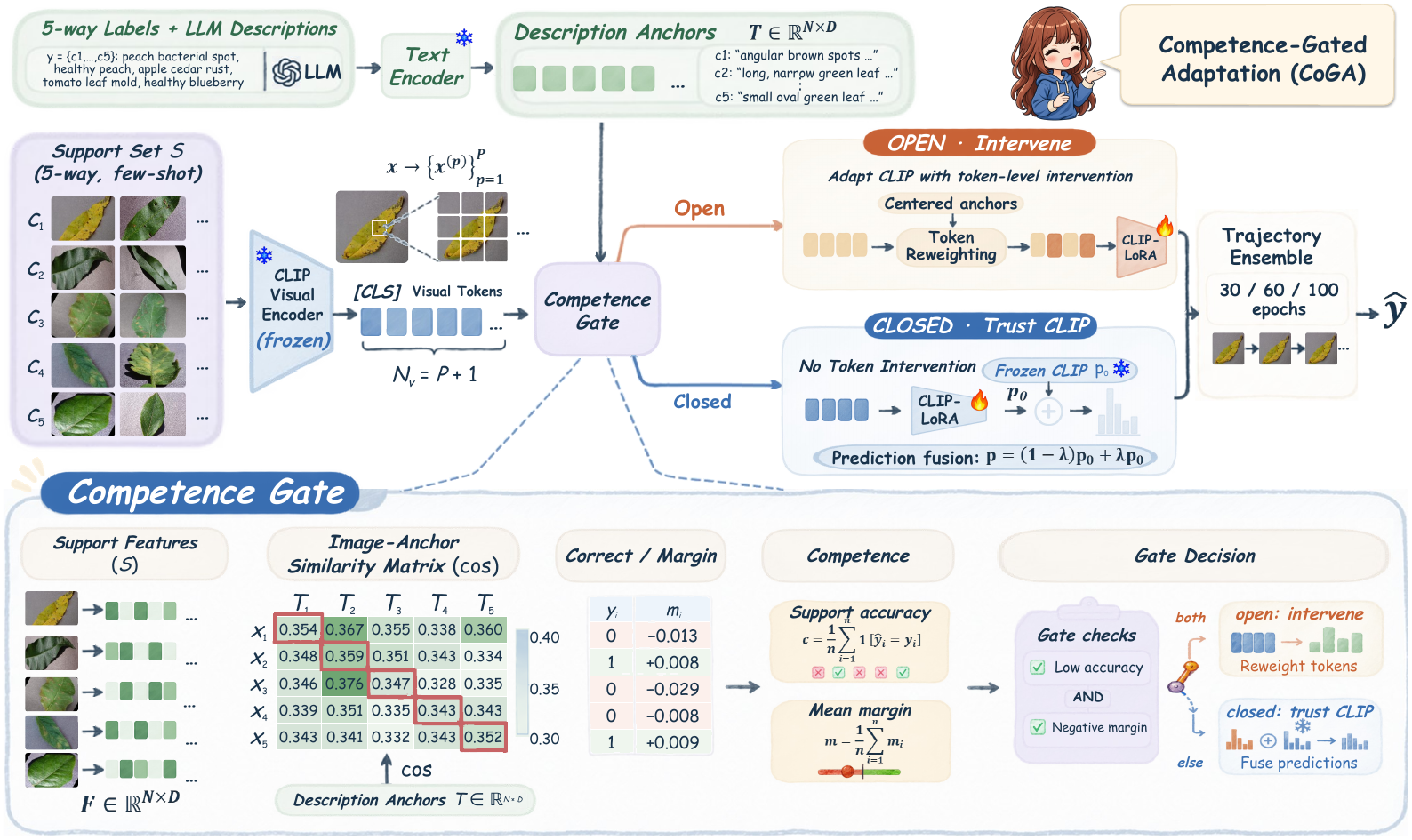}
\captionsetup{skip=3pt}
\caption{Overview of \ours{}. Top: Low frozen CLIP support accuracy $c$ and negative mean margin $m$ via description anchors open the gate for centred-anchor token reweighting. Otherwise, CLIP adapts without reweighting and its predictions are fused with frozen CLIP's. Both branches share a trajectory ensemble. Bottom: Support-only gate example on five leaf classes. Red frames mark true-class scores.}
\label{fig:overview}
\vspace{\dimexpr-2mm+60sp\relax}
\end{figure*}

\subsection{Competence Gate}
\label{sec:gate}

\begin{figure}[t]
  \centering
  \includegraphics[width=\columnwidth]{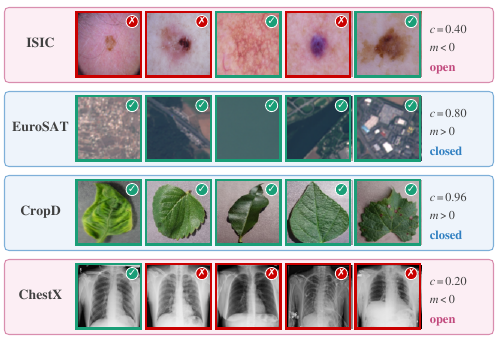}
  \vspace{-6mm}
  \caption{The gate on the first 1-shot episode of each BSCD-FSL domain. Marks: frozen CLIP's prediction on each plain support image; $c$ and $m$ use the five views of each image (\cref{eq:comp,eq:margin}).}
  \label{fig:gatecol}
  \vspace{-3mm}
\end{figure}

Before training on an episode, \ours{} computes the competence $c$ and the margin $m$ of frozen CLIP (\cref{eq:comp,eq:margin}) with one forward pass over the support views, and opens the gate when both indicate that CLIP cannot read the episode,
\begin{equation}
  \mathrm{open}=\mathbbm{1}[c<\tau]\cdot\mathbbm{1}[m<0].
  \label{eq:gate}
\end{equation}
The gate reads frozen CLIP, so the decision is taken before training and is not affected by fitting the support set. Neither cut point is tuned: $c<0.5$ means that frozen CLIP misclassifies most support views, and $m<0$ that its errors are systematic. The same $\tau$ serves every domain, shot, backbone and framework, and the sign change in \cref{fig:flip} agrees with it. \cref{fig:gatecol} shows the gate on the first evaluation episode of each domain.

\subsection{Open Branch: Intervene}
\label{sec:open}

When the gate opens, frozen CLIP's anchors do not separate the classes of the episode (\cref{fig:sim} in the supplement), and the adaptation has to move the features far from the pre-trained solution. The open branch therefore applies our token reweighting (\cref{eq:rule}) at blocks 8 and 10 with $\rho=0.3$ and $\beta=0.5$, and doubles the adapters' learning rate, which raises 5-shot ChestX by 1.30 points (\cref{tab:sens}). The rule ranks tokens against centred anchors $\tilde t_k=\mathrm{norm}\big(t_k-\frac{1}{N}\sum_{k'}t_{k'}\big)$: with descriptions, the anchors share most of their direction (mean pairwise cosine 0.89 on ISIC and ChestX), and centring keeps the ranking on what distinguishes the classes.

\subsection{Closed Branch: Trust CLIP}
\label{sec:closed}

When the gate closes, the support set fails the joint criterion for intervention, and adaptation with only one or five images per class may drift from a useful pre-trained solution. The closed branch therefore preserves frozen CLIP's prediction: it adapts without intervention and, for each snapshot of the trajectory ensemble, fuses the adapted prediction with frozen CLIP's,
\begin{equation}
  p(k\mid x)=(1-\lambda)\,p_\theta(k\mid x)+\lambda\,p_0(k\mid x),
  \label{eq:fuse}
\end{equation}
where $p_\theta$ and $p_0$ are the adapted and frozen predictions, both at CLIP's logit scale of 100, and $\lambda=0.5$. Averaging probabilities keeps frozen CLIP's prediction where the adapted model is uncertain and lets the adapted model correct frozen CLIP where it is confident.

\begin{table*}[t]
\centering
\caption{Comparison on BSCD-FSL (5-way accuracy, \%). ATHA, TIR and \ours{} are evaluated on the same episodes. \emph{Source}: source-data training. \fst{Red}, \snd{green} and \trd{blue} mark the top three in each column.}
\label{tab:main}
\footnotesize
\setlength{\tabcolsep}{3.0pt}
\renewcommand{\arraystretch}{0.94}
\begin{tabularx}{\textwidth}{llllcYYYYY}
\toprule
\rowcolor{headbg}
Shot & Method & Venue & Backbone & Source & ISIC & EuroSAT & CropDiseases & ChestX & Avg. \\
\midrule
\multirow{22}{*}{\rotatebox{90}{1-shot}}
 & \multicolumn{9}{>{\columncolor{setbg}}l}{\textit{Other backbones}} \\
 & StyleAdv-FT~\citep{fu2023styleadv} & CVPR-23 & ViT/DINO & \cmark & 33.99 & 74.93 & 84.11 & 22.92 & 53.99 \\
 & FLoR~\citep{zou2024flatten} & CVPR-24 & ViT/DINO & \cmark & 35.49 & 73.09 & 83.55 & 23.26 & 53.85 \\
 & DAMIM~\citep{ma2025reconstruction} & AAAI-25 & ViT/DINO & \cmark & 36.35 & 73.61 & 83.90 & 23.38 & 54.31 \\
 & CD-CLS~\citep{zou2024closer} & NeurIPS-24 & ViT/DINO & \cmark & 35.56 & 74.97 & 84.53 & 23.39 & 54.62 \\
 & AttnTemp~\citep{zou2024attention} & NeurIPS-24 & ViT/DINO & \cmark & 38.05 & 75.09 & 84.78 & 23.63 & 55.39 \\
 & ReCIT~\citep{yi2025revisitingcontinuityimagetokens} & ICML-25 & ViT/DINO & \cmark & 38.48 & 75.23 & \snd{85.92} & \trd{23.84} & 55.87 \\
 & REAP~\citep{yi2025random} & ICML-25 & ViT/DINO & \cmark & \trd{38.67} & 75.97 & \trd{85.33} & \fst{24.17} & \trd{56.04} \\
 & FN+VDB~\citep{yazdanpanah2022visual} & CVPR-22 & RN18 & -- & 32.96 & 69.67 & 79.68 & 22.64 & 51.24 \\
 & IM-DCL~\citep{xu2024enhancing} & TIP-24 & RN10 & -- & 38.13 & 77.14 & 84.37 & \snd{23.98} & 55.91 \\
 & \multicolumn{9}{>{\columncolor{setbg}}l}{\textit{Source-free CLIP ViT-B/16}} \\
 & StepSTP~\citep{xu2025step} & TPAMI-26 & ViT/CLIP & -- & 32.97 & 70.01 & 84.84 & 22.84 & 52.68 \\
 & CoOp~\citep{zhou2022learning} & IJCV-22 & ViT/CLIP & -- & 32.86 & 72.08 & 80.50 & 21.65 & 51.77 \\
 & Tip-Adapter~\citep{zhang2022tip} & ECCV-22 & ViT/CLIP & -- & 32.68 & 75.44 & 77.15 & 22.24 & 51.87 \\
 & AMU-Tuning~\citep{tang2024amu} & CVPR-24 & ViT/CLIP & -- & 32.29 & 72.24 & 80.20 & 21.56 & 51.57 \\
 & LP++~\citep{huang2024lp++} & CVPR-24 & ViT/CLIP & -- & 33.63 & 73.05 & 81.84 & 21.72 & 52.56 \\
 & LDC~\citep{li2025logits} & CVPR-25 & ViT/CLIP & -- & 33.72 & 74.19 & 83.77 & 22.12 & 53.45 \\
 & MaPLe~\citep{khattak2023maple} & CVPR-23 & ViT/CLIP & -- & 33.38 & 76.05 & 81.78 & 21.09 & 53.07 \\
 & CLIP-LoRA~\citep{zanella2024low} & CVPRW-24 & ViT/CLIP & -- & 35.23 & 81.41 & 85.32 & 21.73 & 55.92 \\
 & CLIP-LoRA + ATHA~\citep{yi2026improving} & ICML-26 & ViT/CLIP & -- & 37.00 & \trd{81.61} & 81.66 & 21.80 & 55.52 \\
 & CLIP-LoRA + TIR~\citep{yi2026addressing} & CVPR-26 & ViT/CLIP & -- & \snd{39.22} & \snd{82.28} & 83.77 & 21.70 & \snd{56.74} \\
\rowcolor{oursbg} & CLIP-LoRA + \ours{} & Ours & ViT/CLIP & -- & \fst{42.49} & \fst{85.29} & \fst{87.78} & 22.67 & \fst{59.56} \\
\midrule
\multirow{22}{*}{\rotatebox{90}{5-shot}}
 & \multicolumn{9}{>{\columncolor{setbg}}l}{\textit{Other backbones}} \\
 & StyleAdv-FT~\citep{fu2023styleadv} & CVPR-23 & ViT/DINO & \cmark & 51.23 & 90.12 & 95.99 & 26.97 & 66.08 \\
 & FLoR~\citep{zou2024flatten} & CVPR-24 & ViT/DINO & \cmark & 53.06 & 90.75 & 96.47 & 27.02 & 66.83 \\
 & DAMIM~\citep{ma2025reconstruction} & AAAI-25 & ViT/DINO & \cmark & 54.86 & 91.18 & 96.34 & 27.82 & 67.55 \\
 & CD-CLS~\citep{zou2024closer} & NeurIPS-24 & ViT/DINO & \cmark & 54.69 & 91.53 & 96.27 & 27.66 & 67.54 \\
 & AttnTemp~\citep{zou2024attention} & NeurIPS-24 & ViT/DINO & \cmark & 54.91 & 90.82 & 96.66 & 28.03 & 67.61 \\
 & ReCIT~\citep{yi2025revisitingcontinuityimagetokens} & ICML-25 & ViT/DINO & \cmark & 54.91 & 91.58 & \trd{96.85} & \snd{28.88} & \snd{68.06} \\
 & REAP~\citep{yi2025random} & ICML-25 & ViT/DINO & \cmark & \trd{55.28} & 91.79 & 96.71 & \trd{28.34} & \trd{68.03} \\
 & FN+VDB~\citep{yazdanpanah2022visual} & CVPR-22 & RN18 & -- & 47.48 & 87.31 & 94.63 & 25.55 & 63.74 \\
 & IM-DCL~\citep{xu2024enhancing} & TIP-24 & RN10 & -- & 52.74 & 89.47 & 95.73 & \fst{28.93} & 66.72 \\
 & \multicolumn{9}{>{\columncolor{setbg}}l}{\textit{Source-free CLIP ViT-B/16}} \\
 & StepSTP~\citep{xu2025step} & TPAMI-26 & ViT/CLIP & -- & 52.12 & 89.40 & 96.01 & 26.36 & 65.97 \\
 & CoOp~\citep{zhou2022learning} & IJCV-22 & ViT/CLIP & -- & 45.78 & 85.88 & 93.31 & 23.35 & 62.08 \\
 & Tip-Adapter~\citep{zhang2022tip} & ECCV-22 & ViT/CLIP & -- & 46.96 & 87.24 & 94.19 & 24.07 & 63.12 \\
 & AMU-Tuning~\citep{tang2024amu} & CVPR-24 & ViT/CLIP & -- & 44.60 & 88.47 & 94.26 & 23.34 & 62.66 \\
 & LP++~\citep{huang2024lp++} & CVPR-24 & ViT/CLIP & -- & 48.49 & 87.48 & 94.47 & 23.89 & 63.58 \\
 & LDC~\citep{li2025logits} & CVPR-25 & ViT/CLIP & -- & 49.70 & 90.82 & 96.71 & 25.89 & 65.78 \\
 & MaPLe~\citep{khattak2023maple} & CVPR-23 & ViT/CLIP & -- & 48.35 & 89.04 & 93.50 & 22.96 & 63.46 \\
 & CLIP-LoRA~\citep{zanella2024low} & CVPRW-24 & ViT/CLIP & -- & 51.10 & \trd{92.52} & 96.21 & 24.13 & 65.99 \\
 & CLIP-LoRA + ATHA~\citep{yi2026improving} & ICML-26 & ViT/CLIP & -- & 54.18 & \snd{92.72} & 96.39 & 25.14 & 67.11 \\
 & CLIP-LoRA + TIR~\citep{yi2026addressing} & CVPR-26 & ViT/CLIP & -- & \snd{55.60} & 92.48 & \snd{97.13} & 25.07 & 67.57 \\
\rowcolor{oursbg} & CLIP-LoRA + \ours{} & Ours & ViT/CLIP & -- & \fst{58.01} & \fst{93.56} & \fst{97.42} & 26.86 & \fst{68.96} \\
\bottomrule
\end{tabularx}
\end{table*}

\subsection{Shared Components}

Three components serve both branches, which adapt from one or a few images per class.

\textbf{Description anchors.} A class name alone says little about the appearance of a skin lesion or a crop disease, and a single support image cannot tell the class's cues from those of the image. For each class we use $M=3$ short descriptions of its visual appearance and encode the prompts ``\{class name\}, \{description\}''; folder names such as \texttt{Apple\_\_\_Apple\_scab} are replaced by readable class names. The anchor is $t_k=\mathrm{norm}\big(\frac{1}{M}\sum_{j=1}^{M}\mathrm{norm}(g(p_{k,j}))\big)$ for the prompts $p_{k,j}$ of class $k$; with the adapted text encoder it serves the classifier and the token ranking, and with the frozen one ($t^0_k$) the gate and $p_0$. \cref{app:desc} of the supplement lists all descriptions.

\textbf{Online crops.} The released code of ATHA and TIR adapts on five fixed views of each support image. We additionally resample a random resized crop (minimum scale 0.5) and a horizontal flip of the support images at every step, so that the adapters cannot fit a fixed set of views.

\textbf{Trajectory ensemble.} We keep the adapter weights at epochs 30, 60 and 100 and predict with $\arg\max_k\sum_e\mathrm{z}(\log p_e)_k$, where $p_e$ is the prediction of snapshot $e$ (after fusion in the closed branch) and $\mathrm{z}(\ell)_k=(\ell_k-\bar\ell)/(\mathrm{sd}(\ell)+\epsilon)$ standardizes a query's scores across the $N$ classes ($\epsilon=10^{-6}$). This uses no query labels and no extra training.

\section{Experiments}
\label{sec:exp}

In this section, we investigate five questions. \textbf{Q1:} Does \ours{} outperform the state of the art? (\cref{sec:q1}) \textbf{Q2:} What does each component contribute? (\cref{sec:q2}) \textbf{Q3:} Does the gate decide well? (\cref{sec:q3}) \textbf{Q4:} Does \ours{} generalize? (\cref{sec:q4}) \textbf{Q5:} Is \ours{} robust and cheap? (\cref{sec:q5}).

\begin{table}[t]
\centering
\caption{Component study of \ours{} (accuracy, \%). We evaluate the cumulative effects of the components in our adaptation framework.}
\label{tab:ablation}
\footnotesize
\setlength{\tabcolsep}{1.6pt}
\begin{tabularx}{\columnwidth}{lYcYYY}
\toprule
\rowcolor{headbg}
Method & ISIC & EuroSAT & CropD & ChestX & Avg. \\
\midrule
\rowcolor{setbg}\multicolumn{6}{l}{\textit{5-shot}} \\
CLIP-LoRA & 51.10 & 92.52 & 96.21 & 24.13 & 65.99 \\
gated adaptation & 55.64 & 92.31 & 96.80 & 24.99 & 67.44 \\
+ online crops & 56.61 & 92.45 & 97.20 & 26.85 & 68.28 \\
+ descriptions & 57.69 & 92.92 & 97.41 & 26.48 & 68.63 \\
+ anchor centring & 57.66 & 92.99 & 97.41 & 26.58 & 68.66 \\
\rowcolor{oursbg}Full \ours{} & \textbf{58.01} & \textbf{93.56} & \textbf{97.42} & \textbf{26.86} & \textbf{68.96} \\
\rowcolor{setbg}\multicolumn{6}{l}{\textit{1-shot}} \\
CLIP-LoRA & 35.23 & 81.41 & 85.32 & 21.73 & 55.92 \\
gated adaptation & 38.26 & 84.45 & 83.44 & 21.90 & 57.01 \\
+ online crops & 39.22 & 83.49 & 85.14 & 22.10 & 57.49 \\
+ descriptions & 42.48 & 84.22 & 87.30 & 22.65 & 59.16 \\
+ anchor centring & 42.01 & 84.39 & 87.52 & 22.66 & 59.14 \\
\rowcolor{oursbg}Full \ours{} & \textbf{42.49} & \textbf{85.29} & \textbf{87.78} & \textbf{22.67} & \textbf{59.56} \\
\bottomrule
\end{tabularx}
\end{table}

\subsection{Experimental Setup}
\label{sec:setup}

\textbf{Datasets and baselines.} Following~\citet{yi2026improving,yi2026addressing}, we evaluate on the four target domains of BSCD-FSL~\citep{guo2020broader}: ISIC2018~\citep{codella2019skin} (dermoscopy), EuroSAT~\citep{helber2019eurosat} (satellite), CropDiseases~\citep{mohanty2016using} (plant leaves) and ChestX~\citep{wang2017chestx} (chest X-ray), and further test on DTD~\citep{cimpoi2014describing}, FGVC-Aircraft~\citep{maji2013fine}, APTOS~\citep{aptos2019-blindness-detection} (retina), Kvasir~\citep{borgli2020hyperkvasir} (endoscopy) and PlantDoc~\citep{singh2020plantdoc} (plants in the field; details in \cref{app:data} of the supplement). We compare with methods that train a DINO ViT on source data, source-free methods with ResNets, and source-free methods with CLIP ViT-B/16, among them CLIP-LoRA~\citep{zanella2024low}, ATHA~\citep{yi2026improving} and TIR~\citep{yi2026addressing}; \cref{tab:main} lists all of them.

\textbf{Implementation and evaluation.} We build on the CLIP-LoRA implementation released with ATHA and TIR and keep its setup: CLIP ViT-B/16, LoRA of rank 16 on both encoders, 100 epochs and the baseline augmentations. Unless otherwise specified, \ours{} uses $\tau=0.5$, $\lambda=0.5$, $\beta=0.5$, $\rho=0.3$, $M=3$ and a crop scale of 0.5 for every domain and shot. Evaluation uses 5-way tasks with 15 query images per class and reports mean accuracy over 800 (1-shot) and 400 (5-shot) episodes, unless otherwise specified. The episodes are shared by methods we run. The branch is selected before adaptation from labeled support views while query labels are held out for evaluation and never enter the routing decision. Baseline settings and evaluation details are given in \cref{app:impl,app:data} of the supplement; paired 95\% intervals use recorded matched episodes.

\subsection{A1: Comparison with the State of the Art}
\label{sec:q1}

\begingroup
\setlength{\parskip}{0pt}

As shown in \cref{tab:main}, \ours{} achieves the best average accuracy in both settings, 59.56\% at 1 shot and 68.96\% at 5 shots, and improves over TIR, the strongest prior CLIP-based method, on every domain at both shots. The lead is particularly large on the domains that frozen CLIP reads, such as CropDiseases at 1 shot (+4.01 points): TIR still intervenes despite informative frozen predictions, while the closed branch preserves these useful predictions through fusion. With five shots, the support set pins down the classes better and this lead narrows, as \cref{sec:theory} predicts. \ours{} improves significantly over the TIR and ATHA reruns in all 16 paired comparisons (\cref{tab:paired} in the supplement) and also exceeds REAP and ReCIT, the strongest methods trained on source data.

\cref{fig:expred} in the supplement shows corrections of both kinds: where the gate closes, TIR moves a residential area to river and a healthy strawberry leaf to a healthy soybean leaf, and the closed branch keeps them in their class. These results show that deciding when to intervene, rather than intervening on every domain, gives the strongest CLIP-based method for SF-CDFSL.

On ChestX, frozen CLIP is close to chance (21\% support accuracy), so the gate opens on every episode, and the open branch improves over TIR by 0.98 (1-shot) and 1.80 (5-shot) points.

\makeatletter
\renewcommand{\cvprsubsection}{\@startsection{subsection}{2}{\z@}{-12pt}{0.01pt}{\elvbf}}
\makeatother
\subsection{A2: Ablation Study}
\label{sec:q2}

\cref{tab:ablation} evaluates components within the gated adaptation pipeline. This cumulative study begins with the two-branch adaptation policy and then adds the shared components, separating its initial gain from later design refinements. With the gate and its two branches, \ours{} reaches 67.44\% at 5 shots and 57.01\% at 1 shot, and \cref{sec:q3} isolates the gate's decision on the same episodes. Online crops add most at 5 shots on ChestX and ISIC (+1.86 and +0.97 points), where the open branch moves the features far from the pre-trained solution and benefits from more diverse views. At 1 shot, descriptions add auxiliary visual cues (\cref{tab:ablation,tab:desccomp}). The initial gains over CLIP-LoRA appear before later refinements (+1.09 at 1 shot and +1.45 at 5 shots), showing the value of the two-branch design itself. Crops and descriptions further refine how the branches adapt and predict. The components jointly enhance competence-guided adaptation. \cref{tab:traj} in the supplement evaluates the trajectory ensemble on the full episode set.

\textbf{Fair comparison.} The lead of \ours{} holds when the baselines share its components. With the same trajectory ensemble applied to both baselines, the stronger one, TIR, reaches 56.99\% (1-shot) and 68.02\% (5-shot), and \ours{} leads it by 2.57 and 0.94 points, respectively. The gains over standard TIR in \cref{tab:main} are 2.82 and 1.39 points. Without descriptions (row ``+ online crops''), \ours{} exceeds both ATHA and TIR in \cref{tab:main}, leading TIR by 0.75 (1-shot) and 0.71 (5-shot) points in average accuracy. The advantage thus holds with the ensemble shared and with the descriptions removed.

\par
\endgroup

\subsection{A3: Analysis of the Gate}
\label{sec:q3}

\begin{table}[t]
\centering
\caption{Switches between the two branches, replayed from forced runs of each (average accuracy, \%). Others: five further domains, 100 episodes each; All: unweighted average over nine domains. $^\dagger$At the gate's per-domain rate. $^\ddagger$Uses query labels.}
\label{tab:switch}
\footnotesize
\setlength{\tabcolsep}{2.4pt}
\begin{tabularx}{\columnwidth}{lYYYYYY}
\toprule
\rowcolor{headbg}
 & \multicolumn{3}{c}{1-shot} & \multicolumn{3}{c}{5-shot} \\
\rowcolor{headbg}
Switch & BSCD & Others & All & BSCD & Others & All \\
\midrule
Always closed & 59.09 & 67.81 & 63.93 & 68.39 & 78.91 & 74.23 \\
Always open & 58.99 & 62.06 & 60.70 & 68.95 & 76.68 & 73.24 \\
Random$^\dagger$ & 59.48 & 68.91 & 64.72 & 68.94 & 79.41 & 74.76 \\
$c<\tau$ only & 59.52 & 68.79 & 64.67 & 68.95 & 79.41 & 74.76 \\
$m<0$ only & 59.49 & 68.94 & 64.74 & 68.95 & 79.41 & 74.76 \\
\rowcolor{oursbg}\ours{} gate & \textbf{59.56} & \textbf{68.96} & \textbf{64.78} & \textbf{68.96} & \textbf{79.42} & \textbf{74.77} \\
\midrule
Domain oracle$^\ddagger$ & 59.57 & 69.07 & 64.85 & 69.00 & 79.46 & 74.81 \\
Episode oracle$^\ddagger$ & 60.86 & 69.57 & 65.70 & 70.03 & 80.15 & 75.65 \\
\bottomrule
\end{tabularx}
\par\vspace{66855sp}
\end{table}
\begin{table}[t]
\centering
\caption{The gate with other interventions in the open branch (accuracy, \%). We compare different intervention rules under the same adaptation framework.}
\label{tab:rule}
\footnotesize
\setlength{\tabcolsep}{2.4pt}
\begin{tabularx}{\columnwidth}{lYYYYY}
\toprule
\rowcolor{headbg}
Open branch & ISIC & EuroSAT & CropD & ChestX & Avg. \\
\midrule
\rowcolor{setbg}\multicolumn{6}{l}{\textit{5-shot}} \\
ATHA & 54.18 & 92.72 & 96.39 & 25.14 & 67.11 \\
Gate + ATHA rule & 55.12 & 93.49 & 97.17 & 26.43 & 68.05 \\
Gate + no rule & 57.07 & 93.54 & 97.41 & 26.63 & 68.66 \\
\rowcolor{oursbg}Full \ours{} & \textbf{58.01} & \textbf{93.56} & \textbf{97.42} & \textbf{26.86} & \textbf{68.96} \\
\rowcolor{setbg}\multicolumn{6}{l}{\textit{1-shot}} \\
ATHA & 37.00 & 81.61 & 81.66 & 21.80 & 55.52 \\
Gate + ATHA rule & 38.07 & 84.97 & 87.25 & 22.08 & 58.09 \\
Gate + no rule & 40.74 & 85.28 & 87.70 & 22.54 & 59.07 \\
\rowcolor{oursbg}Full \ours{} & \textbf{42.49} & \textbf{85.29} & \textbf{87.78} & \textbf{22.67} & \textbf{59.56} \\
\bottomrule
\end{tabularx}
\end{table}

\textbf{Gate versus fixed choices.} \cref{tab:switch} replays different switches on the same episodes. No fixed choice serves all domains: always intervening is the better fixed choice on BSCD-FSL at 5 shots, but costs up to 18.37 points on the five further domains, which frozen CLIP mostly reads. The gate is right in both regimes: over all nine domains, it beats always intervening by 4.08 (1-shot) and 1.53 (5-shot) points and always closing by 0.85 and 0.54, and comes within 0.07 points of a domain-level oracle that uses query labels. Unlike rate-matched random switching, \ours{} selects its intervention directly from each episode's support set without precomputed per-domain opening rates. This lets one fixed competence rule choose between preserving frozen CLIP and updating visual tokens across all tested domains and shots.

\begin{figure}[t]
  \centering
  \includegraphics[width=\columnwidth]{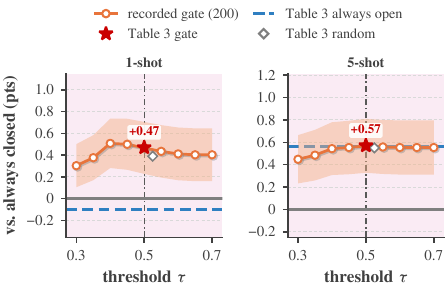}
  \caption{Gate minus always closing versus $\tau$, BSCD-FSL average. Band: paired 95\% confidence interval of the curve. Red star: gate at $\tau=0.5$; dashed: always opening; diamond: random switch. Per domain: \cref{fig:tau} of the supplement.}
  \label{fig:taucol}
\end{figure}

\textbf{Choice of intervention.} The gate is not tied to our reweighting. With ATHA's alignment in the open branch (\cref{tab:rule}), the gate improves ATHA on every domain at both shots, by 2.57 (1-shot) and 0.94 (5-shot) points on average. Our reweighting further improves the no-intervention variant by 0.49 (1-shot) and 0.30 (5-shot) points, particularly on ISIC. This complements the gate's routing, while descriptions, crops, the open branch's learning rate and the closed branch's fusion jointly strengthen the pipeline. The gate is also robust to its threshold: for $\tau$ between 0.4 and 0.7, the average changes by at most 0.11 points (\cref{fig:taucol}). Other interventions can therefore be plugged into the gate.

\subsection{A4: Generalization}
\label{sec:q4}

\begin{table}[t]
\centering
\caption{CoGA on other backbones (LoRA, 1-shot, 200 episodes) and fine-tuning frameworks (ViT-B/16, 5-shot; LoRA rows: 400 episodes; DoRA/SSF: 200). $\Delta$: change in average accuracy.}
\label{tab:general}
\footnotesize
\setlength{\tabcolsep}{1.0pt}
\begin{tabularx}{\columnwidth}{lYcYYYY}
\toprule
\rowcolor{headbg}
Model & ISIC & EuroSAT & CropD & ChestX & Avg. & $\Delta$ \\
\midrule
\rowcolor{setbg}\multicolumn{7}{l}{\textit{Backbones, 1-shot}} \\
SigLIP2 & 42.62 & 83.13 & 87.37 & 21.98 & 58.77 &  \\
\rowcolor{oursbg}\quad + \ours{} & \textbf{45.87} & \textbf{85.91} & \textbf{92.55} & \textbf{22.39} & \textbf{61.68} & \dpos{+2.91} \\
PE-Core & 43.49 & 86.09 & 89.66 & 22.68 & 60.48 &  \\
\rowcolor{oursbg}\quad + \ours{} & \textbf{46.27} & \textbf{90.50} & \textbf{93.04} & \textbf{23.19} & \textbf{63.25} & \dpos{+2.77} \\
ViT-L/14 & 39.40 & 89.05 & 89.63 & 22.46 & 60.14 &  \\
\rowcolor{oursbg}\quad + \ours{} & \textbf{43.84} & \textbf{91.13} & \textbf{91.70} & \textbf{23.05} & \textbf{62.43} & \dpos{+2.29} \\
\midrule
\rowcolor{setbg}\multicolumn{7}{l}{\textit{Fine-tuning frameworks, 5-shot}} \\
LoRA & 51.10 & 92.52 & 96.21 & 24.13 & 65.99 &  \\
\rowcolor{oursbg}\quad + \ours{} & \textbf{58.01} & \textbf{93.56} & \textbf{97.42} & \textbf{26.86} & \textbf{68.96} & \dpos{+2.97} \\
DoRA & 54.02 & 92.82 & 96.99 & 24.91 & 67.19 &  \\
\rowcolor{oursbg}\quad + \ours{} & \textbf{57.58} & \textbf{93.55} & \textbf{97.36} & \textbf{26.75} & \textbf{68.81} & \dpos{+1.63} \\
SSF & 52.24 & 93.15 & 95.96 & 23.91 & 66.31 &  \\
\rowcolor{oursbg}\quad + \ours{} & \textbf{57.00} & \textbf{93.49} & \textbf{97.01} & \textbf{27.23} & \textbf{68.68} & \dpos{+2.37} \\
\bottomrule
\end{tabularx}
\end{table}
\begin{table}[t]
\centering
\caption{Beyond 5-way BSCD-FSL (accuracy, \%, 100 episodes per cell). 7-/10-way: 1-shot, averaged over BSCD-FSL / over EuroSAT, CropDiseases, DTD and FGVC-Aircraft; others: mean of 1- and 5-shot. $c$: frozen CLIP's average support accuracy. All rows: our runs on the same episodes, TIR at 7/10 ways with its released rule for more than five classes. Details: \cref{tab:more} of the supplement.}
\label{tab:beyond}
\footnotesize
\setlength{\tabcolsep}{1.0pt}
\begin{tabularx}{\columnwidth}{lYYYYYYc}
\toprule
\rowcolor{headbg}
Method & 7-way & 10-way & DTD & Aircraft & APTOS & Kvasir & PlantDoc \\
\rowcolor{setbg}
\textit{frozen $c$} & -- & -- & 0.82 & 0.80 & 0.21 & 0.24 & 0.75 \\
\midrule
CLIP-LoRA & 52.27 & 71.85 & 85.28 & 79.21 & 41.05 & 71.80 & 70.58 \\
+ ATHA & 51.05 & 68.74 & 83.48 & 74.45 & 40.44 & 70.97 & 67.53 \\
+ TIR & 51.89 & 70.55 & 83.39 & 77.01 & 42.64 & 72.21 & 68.40 \\
\rowcolor{oursbg}+ \ours{} & \textbf{54.58} & \textbf{76.53} & \textbf{89.23} & \textbf{84.35} & \textbf{44.33} & \textbf{75.84} & \textbf{77.19} \\
\bottomrule
\end{tabularx}
\end{table}

\textbf{Backbones and fine-tuning frameworks.} Following~\citet{yi2026improving,yi2026addressing}, we apply \ours{} to SigLIP2-B/16~\citep{tschannen2025siglip}, PE-Core-B/16~\citep{bolya2026perception} and CLIP ViT-L/14 at 1 shot and compare against our own runs of these baselines on the same episodes, which are 3.9 to 9.3 points stronger than the averages reported in prior work. \ours{} improves every domain on every backbone, raising the average by 2.29 to 2.91 points (\cref{tab:general}), and without retuning $\tau$ the gate opens on most ISIC and ChestX episodes and stays closed on nearly all EuroSAT and CropDiseases episodes. With DoRA~\citep{liu2024dora} or SSF~\citep{lian2022scaling} in place of LoRA, \ours{} again improves every domain and raises the 5-shot average by 1.63 and 2.37 points. With LoRA, \ours{} reaches 68.96\% at 5 shots, 2.97 points above CLIP-LoRA. The gate therefore transfers across backbones and adaptation frameworks.

\textbf{More ways and more domains.} \cref{tab:beyond} tests the analysis beyond BSCD-FSL. These domains cover both sides of the competence split, testing the gate where frozen predictions are useful and active intervention is needed. Frozen CLIP reads DTD, FGVC-Aircraft and PlantDoc ($c\geq0.74$), and there ATHA and TIR fall below CLIP-LoRA, by up to 7.61 points, as \cref{sec:theory} predicts; the gate stays closed on almost all of their episodes, and \ours{} improves on CLIP-LoRA instead, by 3.95 to 6.61 points. APTOS and Kvasir ($c\leq0.25$) open the gate, and \ours{} is again the best method, 3.28 and 4.04 points above CLIP-LoRA. With seven and ten classes, ATHA and TIR again fall below CLIP-LoRA, while \ours{} is the best method on every domain, 2.31 and 4.68 points above CLIP-LoRA on average. The competence signal thus predicts where intervening helps beyond the benchmark used to design the method.

\subsection{A5: Sensitivity and Cost}
\label{sec:q5}

Varied one at a time (\cref{tab:sens,fig:tau} in the supplement), the open-branch learning-rate multiplier, the crop scale, and the number of descriptions keep the 5-shot average within 0.45 points. At 5 shots, $\beta$ and $\rho$ peak at their default settings; their 1-shot differences remain modest. The fusion weight $\lambda$ is robust between 0 and 0.5, where the more accurate adapted prediction keeps at least half of the weight; from 0.75 on, frozen CLIP's zero-shot prediction, 26.74 points less accurate on EuroSAT and CropDiseases ($\lambda=1$), takes over. The defaults thus hold across domains without tuning.

As for cost, \ours{} adds no trainable parameters, two forward passes of frozen CLIP (support views and queries) and about 0.5 GB of memory to CLIP-LoRA, whereas ATHA needs 4.5 times its memory at 1 shot (\cref{tab:cost} in the supplement). A 5-shot episode takes about as long as with CLIP-LoRA, so the gate comes at almost no extra cost.


\section{Conclusion}
\label{sec:conclusion}
We study when visual-token intervention benefits CLIP in SF-CDFSL. Experiments and a bias--variance analysis reveal that intervention helps when frozen CLIP struggles but can hurt when its predictions are reliable. Support-set accuracy and margin provide a practical signal for this choice. Guided by this insight, \ours{} decides whether to intervene or retain frozen CLIP's predictions, without query labels or extra trainable parameters. On BSCD-FSL, \ours{} achieves the highest mean accuracy at 1 and 5 shots (59.56\% and 68.96\%), with gains across interventions, backbones, fine-tuning frameworks, and target domains. This favors competence-guided adaptation over uniform intervention.

\section*{LLM Usage Statement}

OpenAI's ChatGPT (GPT) was used solely for language polishing of the manuscript, including grammar, wording, and readability. The authors reviewed all suggested text and take full responsibility for the research design, implementation, data analysis, figures, experimental results, and all scientific claims presented in the paper.

{
    \small
    \bibliographystyle{ieeenat_fullname}
    \bibliography{refs}
}

\end{document}


\onecolumn
\raggedbottom   
\begin{center}
  {\Large\textbf{Intervene or Trust CLIP? Competence-Gated Adaptation for\\Source-Free Cross-Domain Few-Shot Learning}}\\[0.6em]
  {\large Supplementary Material}
\end{center}
\vspace{0.5em}
\appendix
\renewcommand{\thefigure}{S\arabic{figure}}\renewcommand{\theHfigure}{S\arabic{figure}}
\renewcommand{\thetable}{S\arabic{table}}\renewcommand{\theHtable}{S\arabic{table}}
\renewcommand{\thealgorithm}{S\arabic{algorithm}}\renewcommand{\theHalgorithm}{S\arabic{algorithm}}
\renewcommand{\theequation}{S\arabic{equation}}\renewcommand{\theHequation}{S\arabic{equation}}
\renewcommand{\theproposition}{S\arabic{proposition}}\renewcommand{\theHproposition}{S\arabic{proposition}}

\section{Algorithm}
\label{app:alg}

\begin{algorithm}[h]
\caption{\ours{} on one episode ($\theta_e$: the adapted parameters after epoch $e$)}
\label{alg:coga}
\begin{algorithmic}[1]
\small
\REQUIRE support set $\mathcal{S}$ and its views $\mathcal{S}_{\mathrm{aug}}$, queries $\mathcal{Q}$, frozen CLIP $(f_0,g_0)$, frozen description anchors $\{t^0_k\}$, $\tau$, $\lambda$, base LoRA learning rate $\eta$
\STATE compute $c$ and $m$ on $\mathcal{S}_{\mathrm{aug}}$ with frozen CLIP \hfill Eq.~(\ref{eq:comp}--\ref{eq:margin})
\IF{$c<\tau$ \AND $m<0$}
  \STATE train LoRA on $\mathcal{S}$ with our centred-anchor reweighting, Eq.~(\ref{eq:rule}), and lr $2\eta$
  \STATE $p_e\leftarrow p_{\theta_e}$ for snapshots $e\in\{30,60,100\}$ \hfill (open)
\ELSE
  \STATE train LoRA on $\mathcal{S}$ with lr $\eta$
  \STATE $p_e\leftarrow(1-\lambda)\,p_{\theta_e}+\lambda\,p_0$ for $e\in\{30,60,100\}$ \hfill (closed)
\ENDIF
\STATE \textbf{return} $\hat y(x)=\arg\max_k\sum_{e}\mathrm{z}(\log p_e(x))_k$ for $x\in\mathcal{Q}$
\end{algorithmic}
\end{algorithm}

\section{Theory}
\label{app:proof}

\textbf{Setting.} We use the model of \cref{sec:theory}: class $k$ has $d$-dimensional features $f(x)=\mu_k+\xi$ with $\xi\sim\mathcal{N}(0,\sigma^2 I_d)$, frozen CLIP's anchor is $t^0_k=\mu_k+b_k$ with $\|b_k\|=\delta$, and trusting frozen CLIP to a degree $\gamma$ gives the class weight $w_k(\gamma)=(1-\gamma)\hat\mu_k+\gamma\,t^0_k$. For frozen CLIP's competence we further assume that the centres $\mu_k$ have pairwise distance $\Delta$ and that the biases are orthogonal to the centres and to each other. The model measures the margin with squared distances, $m_{\mathrm{dist}}$, in place of the cosines of \cref{eq:margin}; for unit-norm features and anchors, as in CLIP, $\|x-t\|^2=2-2\cos(x,t)$, so nearest-anchor and cosine classification coincide and $m_{\mathrm{dist}}=2m$. The Gaussian model is not unit-norm, and normalizing its features would change the derivation, so this correspondence motivates the use of $c$ and $m$ rather than proving that the expected cosine margin is monotone in $\delta$. The statements below hold for fixed $\Delta$, $\sigma$, $d$ and $N$.

\begin{proposition}
\label{prop:comp}
Let frozen CLIP classify a support image by its nearest anchor. Then
\begin{align}
  \mathbb{E}[c]&=F_N\big(s(\delta)\big),\quad s(\delta)=\frac{\Delta^2}{2\sigma\sqrt{\Delta^2+2\delta^2}},\label{eq:compth}\\
  \mathbb{E}[m_{\mathrm{dist}}]&=\Delta^2-2\sigma e_N\sqrt{\Delta^2+2\delta^2},\label{eq:marginth}
\end{align}
where $F_N$ and $e_N$ are the distribution function and the mean of the maximum of $N-1$ standard Gaussians with pairwise correlation $1/2$. Both $\mathbb{E}[c]$ and $\mathbb{E}[m_{\mathrm{dist}}]$ decrease strictly in $\delta$ (for $N\geq3$), and the open branch has the lower error exactly when
\begin{equation}
  \mathbb{E}[c]<c^\star_K=F_N\bigg(\frac{\Delta^2}{2\sigma\sqrt{\Delta^2+2\kappa d\sigma^2/K}}\bigg),
  \label{eq:thr}
\end{equation}
a threshold that grows with $K$.
\end{proposition}

\textbf{Proof of Proposition~\ref{prop:shrink}.} By the definition of $w_k(\gamma)$, $w_k(\gamma)-\mu_k=(1-\gamma)(\hat\mu_k-\mu_k)+\gamma\,b_k$. The estimation error $\hat\mu_k-\mu_k$ is the mean of $K$ independent draws of $\xi$, so it has zero mean and $\mathbb{E}\|\hat\mu_k-\mu_k\|^2=d\sigma^2/K=v$, and the cross term vanishes in expectation:
\begin{equation*}
  R(\gamma)=(1-\gamma)^2v+2\gamma(1-\gamma)\,\mathbb{E}\langle\hat\mu_k-\mu_k,b_k\rangle+\gamma^2\delta^2=(1-\gamma)^2v+\gamma^2\delta^2 .
\end{equation*}
$R$ is a convex quadratic with $R'(\gamma)=-2(1-\gamma)v+2\gamma\delta^2$, which vanishes at $\gamma^\star=v/(v+\delta^2)$. For $\gamma_o<\gamma_c$,
\begin{equation*}
  R(\gamma_o)-R(\gamma_c)=(\gamma_c-\gamma_o)\big[(2-\gamma_o-\gamma_c)\,v-(\gamma_o+\gamma_c)\,\delta^2\big],
\end{equation*}
which is negative if and only if $\delta^2>\kappa v$, that is \cref{eq:cond}.

\textbf{Proof of Proposition~\ref{prop:comp}.} Take a support feature $x=\mu_y+\xi$ of class $y$ and, for $k\neq y$, the squared-distance margin $Z_k=\|x-t^0_k\|^2-\|x-t^0_y\|^2$. Expanding with $t^0_k=\mu_k+b_k$, $\|\mu_y-\mu_k\|=\Delta$, $\|b_k\|=\delta$ and the orthogonality of the biases to the centres and to each other,
\begin{equation*}
  Z_k=\Delta^2+2\langle\xi,a_k\rangle,\qquad a_k=\mu_y-\mu_k+b_y-b_k .
\end{equation*}
The centres form a regular simplex, so $\langle\mu_y-\mu_k,\mu_y-\mu_j\rangle=\Delta^2/2$ for $j\neq k$, and
\begin{equation*}
  \|a_k\|^2=\Delta^2+2\delta^2,\qquad \langle a_k,a_j\rangle=\tfrac{1}{2}\Delta^2+\delta^2 .
\end{equation*}
The margins $(Z_k)_{k\neq y}$ are therefore jointly Gaussian with mean $\Delta^2$, variance $4\sigma^2(\Delta^2+2\delta^2)$ and pairwise correlation $1/2$ for every $\delta$. Writing $Z_k=\Delta^2-2\sigma\sqrt{\Delta^2+2\delta^2}\,G_k$ with standard Gaussians $G_k$ of pairwise correlation $1/2$, the image is classified correctly if and only if $\min_k Z_k>0$, that is $\max_k G_k<s(\delta)$, which has probability $F_N(s(\delta))$; averaging over the support set gives \cref{eq:compth}. The margin to the strongest wrong class is $\min_k Z_k=\Delta^2-2\sigma\sqrt{\Delta^2+2\delta^2}\,\max_k G_k$, whose mean is \cref{eq:marginth}. Since $s(\delta)$ decreases strictly in $\delta$ and $F_N$ increases strictly, $\mathbb{E}[c]$ decreases strictly; since $e_N>0$ for $N\geq3$, so does $\mathbb{E}[m_{\mathrm{dist}}]$.

\textbf{Proof of \cref{eq:thr}.} By \cref{eq:cond}, the open branch has the lower error if and only if $\delta>\sqrt{\kappa v}$. As $\mathbb{E}[c]=F_N(s(\delta))$ is strictly decreasing in $\delta$, this holds if and only if $\mathbb{E}[c]<F_N\big(s(\sqrt{\kappa v})\big)=c^\star_K$. As $v=d\sigma^2/K$ falls with $K$, $c^\star_K$ grows with $K$.

\section{Implementation Details}
\label{app:impl}

\textbf{Adaptation.} All methods we run share the CLIP-LoRA implementation released with ATHA and TIR. LoRA with rank 16 and $\alpha=8$ is applied to the query, key and value projections of every block of both encoders and trained with AdamW at a learning rate of $2\times10^{-4}$; a prompt context of four tokens, initialised from ``a photo of a'' with the class token at the end, is trained with SGD (learning rate $2\times10^{-3}$, momentum 0.9, weight decay $5\times10^{-4}$). Each episode is trained for 100 epochs over the full support set. The released loader turns every support image into five views (the plain image; rotation, colour jitter and flip together; and each of the three alone), and \ours{} additionally resamples a random resized crop with minimum scale 0.5 and a horizontal flip of the support views at every step.

\textbf{\ours{}.} The gate reads frozen CLIP on the five support views with the description anchors and opens when $c<0.5$ and $m<0$. The open branch applies our updated reweighting (\cref{eq:rule}) at blocks 8 and 10 (blocks 16 and 20 for the 24-block ViT-L/14) with $\rho=0.3$, $\beta=0.5$ and $s_0=3$ on the $\lfloor\rho L\rfloor$ most similar tokens per class, ranks tokens against centred anchors, and doubles the LoRA learning rate. Our open branch keeps $s_0=3$ and $\beta=0.5$ for 7-way and 10-way tasks, so a token selected by more than five classes receives a negative weight; the TIR baseline at these sizes uses the rule its released code gives for more than five classes, which leaves such tokens unchanged. The closed branch fuses the adapted and frozen probabilities with $\lambda=0.5$ at logit scale 100. Both branches keep snapshots at epochs 30, 60 and 100 for the trajectory ensemble. Unless otherwise specified, these defaults are used across domains, shots, backbones and fine-tuning frameworks.

\textbf{Baselines.} We use CLIP-LoRA-Vision~\citep{zanella2024low} as the BSCD-FSL baseline in Tables~\ref{tab:main}, \ref{tab:ablation} and \ref{tab:general}. ATHA and TIR are run from their official code on the same episodes and with the hyperparameters of their papers (ATHA: 10\% of the tokens added and removed; TIR: $\rho=0.3$, $\beta=0.5$ at blocks 8 and 10), 100 epochs and the same five support views; for ATHA at 5 shots, the support step is split into two accumulated halves to fit the device memory, which leaves the update unchanged. For the other backbones and frameworks, the baseline is the same pipeline without the gate and the shared components.

\textbf{Evaluation.} Experiments follow the default settings and episodic evaluation protocol in \cref{sec:setup}, unless otherwise specified. Episodes are sampled once per domain and shot with a fixed seed and shared by methods we run. Confidence intervals are 95\% intervals of paired differences on matched episodes. Experiments run on Ascend 910B NPUs with 64 GB of memory.

\section{Datasets}
\label{app:data}

\begin{table}[H]
\centering
\caption{Target domains. All images are resized to $224\times224$ at the model input.}
\label{tab:data}
\small
\begin{tabular}{llrr}
\toprule
\rowcolor{headbg}
Dataset & Domain & Classes & Images \\
\midrule
ISIC2018~\citep{codella2019skin} & dermoscopy & 7 & 10{,}015 \\
EuroSAT~\citep{helber2019eurosat} & satellite & 10 & 27{,}000 \\
CropDiseases~\citep{mohanty2016using} & plant leaves & 38 & 70{,}295 \\
ChestX~\citep{wang2017chestx} & chest X-ray & 7 & 25{,}848 \\
\midrule
DTD~\citep{cimpoi2014describing} & textures & 47 & 5{,}640 \\
FGVC-Aircraft~\citep{maji2013fine} & aircraft variants & 100 & 6{,}667 \\
APTOS~\citep{aptos2019-blindness-detection} & retinal fundus (DR grade) & 5 & 2{,}329 \\
Kvasir~\citep{borgli2020hyperkvasir} & GI endoscopy & 20 & 10{,}636 \\
PlantDoc~\citep{singh2020plantdoc} & plant disease in the field & 27 & 2{,}463 \\
\bottomrule
\end{tabular}
\end{table}

\textbf{BSCD-FSL.} The four target domains of BSCD-FSL~\citep{guo2020broader} move away from natural images in increasing steps. \emph{CropDiseases}~\citep{mohanty2016using} contains close-up photographs of plant leaves in 38 classes of crop and disease, taken on a uniform background; the classes differ in fine lesion patterns on otherwise similar leaves. \emph{EuroSAT}~\citep{helber2019eurosat} contains Sentinel-2 satellite images of 10 land-use and land-cover classes seen from above, where texture and colour replace object shape. \emph{ISIC2018}~\citep{codella2019skin} contains dermoscopic images of 7 skin-lesion types, which differ in pigment network, border and colour structure under magnification. \emph{ChestX}~\citep{wang2017chestx} contains frontal chest X-rays of 7 thoracic conditions; the images are monochrome, the findings are small and subtle, and it is the domain farthest from CLIP's pre-training data.

\begin{figure}[tbp]
  \centering
  \includegraphics[width=\textwidth]{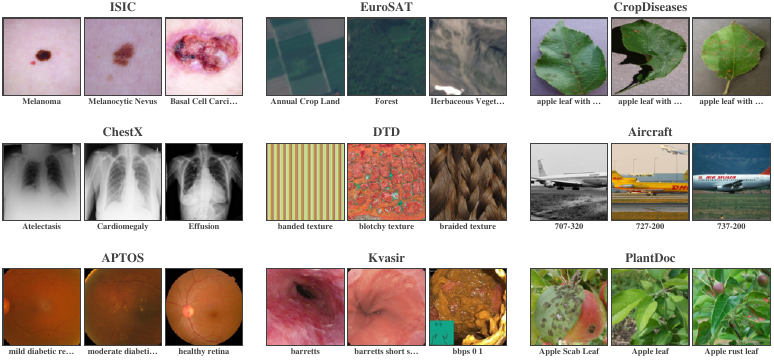}
  \caption{The nine target domains: the first image of each of the first three classes, with the class names the model reads.}
  \label{fig:exdata}
\end{figure}

\textbf{Further domains.} DTD~\citep{cimpoi2014describing} (47 texture classes) and FGVC-Aircraft~\citep{maji2013fine} (100 aircraft variants) are domains that CLIP reads well but whose classes are fine-grained. APTOS~\citep{aptos2019-blindness-detection} grades diabetic retinopathy on retinal fundus photographs in five levels, Kvasir~\citep{borgli2020hyperkvasir} contains gastrointestinal endoscopy images of anatomical landmarks, findings and procedures, and PlantDoc~\citep{singh2020plantdoc} contains plant-disease photographs taken in the field, with cluttered backgrounds instead of the uniform ones of CropDiseases. For these five domains, every labeled image of the release is a candidate (APTOS: the labeled training images of the competition; Kvasir: the labeled images of HyperKvasir), classes with fewer than 25 images are removed (a 5-shot episode draws 20 images per class), and class names are taken from the release with underscores and dashes replaced by spaces, the APTOS grades spelled out (for example, ``proliferative diabetic retinopathy'' for grade 4). \cref{tab:data} lists the resulting sizes and \cref{fig:exdata} shows one image of three classes of each domain.

\section{Additional Results}
\label{app:more}

\subsection{Paired Comparison with ATHA and TIR}

\cref{tab:paired} gives the paired gain of \ours{} over the ATHA and TIR reruns on every reported episode. All 16 gains are significant, and \ours{} is at least as accurate as the baseline on 61\% to 85\% of the episodes of every domain.

\begin{table}[H]
\centering
\caption{Paired gain of \ours{} over the ATHA and TIR reruns (accuracy points, mean $\pm$ 95\% confidence interval, all reported episodes). Gains use unrounded episode accuracies. Second line: share of episodes on which \ours{} is at least as accurate.}
\label{tab:paired}
\small
\setlength{\tabcolsep}{5pt}
\begin{tabular}{llcccc}
\toprule
\rowcolor{headbg}
Shot & Against & ISIC & EuroSAT & CropD & ChestX \\
\midrule
\multirow{4}{*}{1} & ATHA & \dpos{+5.49}$\pm$0.51 & \dpos{+3.67}$\pm$0.51 & \dpos{+6.12}$\pm$0.50 & \dpos{+0.88}$\pm$0.35 \\
 & & 81\% & 74\% & 85\% & 61\% \\
 & TIR & \dpos{+3.27}$\pm$0.45 & \dpos{+3.00}$\pm$0.45 & \dpos{+4.01}$\pm$0.46 & \dpos{+0.98}$\pm$0.34 \\
 & & 72\% & 73\% & 77\% & 61\% \\
\midrule
\multirow{4}{*}{5} & ATHA & \dpos{+3.84}$\pm$0.57 & \dpos{+0.84}$\pm$0.33 & \dpos{+1.03}$\pm$0.28 & \dpos{+1.72}$\pm$0.58 \\
 & & 79\% & 68\% & 80\% & 66\% \\
 & TIR & \dpos{+2.41}$\pm$0.53 & \dpos{+1.07}$\pm$0.34 & \dpos{+0.29}$\pm$0.23 & \dpos{+1.80}$\pm$0.56 \\
 & & 72\% & 69\% & 73\% & 68\% \\
\bottomrule
\end{tabular}
\end{table}

\FloatBarrier
\subsection{The Competence Effect on Nine Domains}

\begin{figure}[H]
  \centering
  \includegraphics[width=0.62\textwidth]{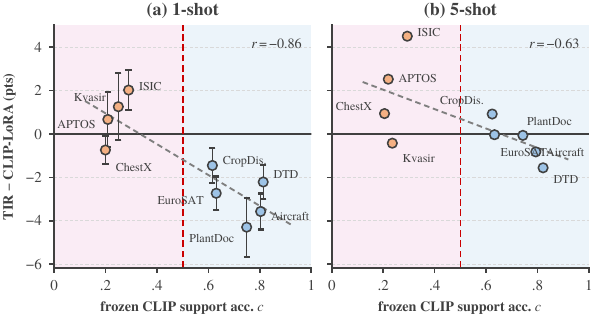}
  \caption{A uniform intervention helps where frozen CLIP cannot read the domain and hurts where it can. One point per domain: average support accuracy $c$ and TIR gain over CLIP-LoRA. BSCD-FSL: paired runs on the first 200 episodes at 1 shot; 400 TIR episodes against the CLIP-LoRA row in \cref{tab:main} at 5 shots. Other domains: 100 matched episodes per shot. Error bars: paired 95\% intervals at 1 shot; 5-shot panels show means only.}
  \label{fig:domapp}
\end{figure}

\begin{figure}[H]
  \centering
  \includegraphics[width=0.94\textwidth]{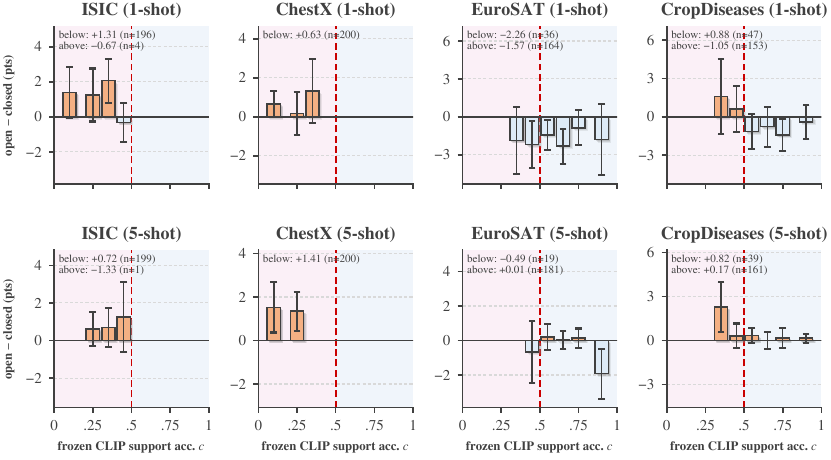}
  \caption{\cref{fig:flip} per domain (first 200 episodes; bins with fewer than 8 episodes omitted). Labels: mean difference and number of episodes on each side of $\tau=0.5$.}
  \label{fig:flipdom}
\end{figure}

With one point per domain, \cref{fig:domapp} extends \cref{sec:analysis} to nine target domains using the paired diagnostic runs at 1 shot. Accuracy and gain are strongly anti-correlated ($r=-0.86$ at 1 shot; $r=-0.63$ at 5 shots). In the 1-shot paired diagnostic, TIR lowers CLIP-LoRA on all five domains that frozen CLIP reads ($c>0.5$), by 1.45 (CropDiseases) to 4.29 (PlantDoc) points, and raises it on ISIC, APTOS and Kvasir, which frozen CLIP cannot read. \cref{fig:flipdom} splits \cref{fig:flip} by domain: the intervention gains on ISIC and ChestX at both shots, and costs accuracy on EuroSAT and CropDiseases at 1 shot, so the sign of the effect follows the domain, which the gate reads from one support set.

\FloatBarrier
\subsection{More Ways, More Domains, Backbones and Cost}

\cref{tab:more} gives the 7-way and 10-way episodes and the five further domains of \cref{sec:q4}: \ours{} is the best method in all 18 settings.

\begin{table}[H]
\centering
\caption{Beyond 5-way BSCD-FSL (accuracy, \%, 100 episodes per cell). Top: 7-way and 10-way 1-shot (10-way on the domains with at least ten classes); TIR with the rule its released code gives for more than five classes. Middle: DTD and FGVC-Aircraft. Bottom: APTOS (diabetic retinopathy), Kvasir (endoscopy) and PlantDoc (plant disease in the field). CLIP-LoRA, ATHA and TIR are our runs on the same episodes.}
\label{tab:more}
\footnotesize
\setlength{\tabcolsep}{1.6pt}
\begin{tabularx}{\columnwidth}{lYcYYY}
\toprule
\rowcolor{headbg}
7-way 1-shot & ISIC & EuroSAT & CropD & ChestX & Avg. \\
\midrule
CLIP-LoRA & 31.02 & 79.69 & 82.46 & 15.90 & 52.27 \\
+ ATHA & 30.36 & 77.24 & 80.30 & 16.30 & 51.05 \\
+ TIR & 30.64 & 78.83 & 82.21 & 15.87 & 51.89 \\
\rowcolor{oursbg}+ \ours{} & \textbf{34.72} & \textbf{81.35} & \textbf{85.82} & \textbf{16.41} & \textbf{54.58} \\
\midrule
\rowcolor{headbg}
10-way 1-shot & EuroSAT & CropD & DTD & Aircraft & Avg. \\
\midrule
CLIP-LoRA & 75.31 & 77.44 & 74.84 & 59.80 & 71.85 \\
+ ATHA & 73.53 & 75.31 & 71.91 & 54.19 & 68.74 \\
+ TIR & 74.17 & 77.05 & 73.97 & 57.01 & 70.55 \\
\rowcolor{oursbg}+ \ours{} & \textbf{76.77} & \textbf{79.79} & \textbf{80.43} & \textbf{69.13} & \textbf{76.53} \\
\midrule
\rowcolor{headbg}
Extra domains & \multicolumn{2}{c}{DTD} & \multicolumn{2}{c}{Aircraft} & \\
\rowcolor{headbg}
 & 1-shot & 5-shot & 1-shot & 5-shot & Avg. \\
\midrule
CLIP-LoRA & 83.24 & 87.32 & 72.45 & 85.97 & 82.25 \\
+ ATHA & 79.87 & 87.09 & 64.84 & 84.07 & 78.97 \\
+ TIR & 81.03 & 85.76 & 68.88 & 85.15 & 80.20 \\
\rowcolor{oursbg}+ \ours{} & \textbf{87.27} & \textbf{91.20} & \textbf{79.28} & \textbf{89.43} & \textbf{86.79} \\
\bottomrule
\end{tabularx}
\par\vspace{2mm}
\begin{tabularx}{\columnwidth}{lYYYYYYY}
\toprule
\rowcolor{headbg}
Extra domains & \multicolumn{2}{c}{APTOS} & \multicolumn{2}{c}{Kvasir} & \multicolumn{2}{c}{PlantDoc} & \\
\rowcolor{headbg}
 & 1-shot & 5-shot & 1-shot & 5-shot & 1-shot & 5-shot & Avg. \\
\midrule
CLIP-LoRA & 34.96 & 47.13 & 62.15 & 81.45 & 64.33 & 76.83 & 61.14 \\
+ ATHA & 33.29 & 47.59 & 62.24 & 79.71 & 58.40 & 76.67 & 59.65 \\
+ TIR & 35.63 & 49.65 & 63.40 & 81.03 & 60.04 & 76.76 & 61.08 \\
\rowcolor{oursbg}+ \ours{} & \textbf{38.07} & \textbf{50.59} & \textbf{68.57} & \textbf{83.11} & \textbf{71.61} & \textbf{82.76} & \textbf{65.78} \\
\bottomrule
\end{tabularx}
\end{table}

\begin{table}[H]
\centering
\caption{Cost per episode on ISIC (one Ascend 910B NPU, no other job): trainable parameters, training time (mean of five episodes after the first) and peak memory.}
\label{tab:cost}
\footnotesize
\setlength{\tabcolsep}{3.0pt}
\begin{tabular}{lccccc}
\toprule
\rowcolor{headbg}
 & & \multicolumn{2}{c}{Time (s)} & \multicolumn{2}{c}{Memory (GB)} \\
\rowcolor{headbg}
Method & Params & 1-shot & 5-shot & 1-shot & 5-shot \\
\midrule
CLIP-LoRA & 1.48M & 20.9 & 66.3 & 5.4 & 18.9 \\
+ ATHA & 1.48M & 29.9 & 83.7 & 24.1 & 38.9 \\
+ TIR & 1.48M & 24.6 & 69.3 & 5.4 & 19.1 \\
\rowcolor{oursbg}+ \ours{} & 1.48M & 32.1 & 61.9 & 5.9 & 19.4 \\
\bottomrule
\end{tabular}
\end{table}

\cref{tab:pe5} completes the backbone study of \cref{sec:q4} at 5 shots: on PE-Core-B/16, \ours{} improves every domain and the average by 1.22 points.

\begin{table}[H]
\centering
\caption{\ours{} on PE-Core-B/16 at 5 shots (accuracy, \%, 200 episodes per domain; baseline: our run on the same episodes).}
\label{tab:pe5}
\small
\setlength{\tabcolsep}{5pt}
\begin{tabular}{lccccc}
\toprule
\rowcolor{headbg}
Method & ISIC & EuroSAT & CropD & ChestX & Avg. \\
\midrule
PE-Core + LoRA & 58.69 & 94.41 & 97.46 & 26.96 & 69.38 \\
\rowcolor{oursbg}+ \ours{} & \textbf{60.41} & \textbf{95.73} & \textbf{98.02} & \textbf{28.24} & \textbf{70.60} \\
\bottomrule
\end{tabular}
\end{table}

\FloatBarrier
\subsection{Competence across Domains}

\begin{figure}[H]
  \centering
  \includegraphics[width=0.55\textwidth]{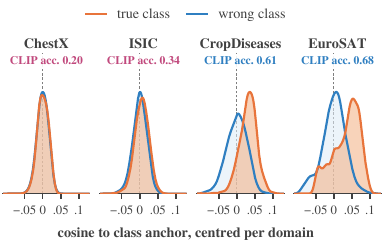}
  \caption{Frozen CLIP's cosine to the true and wrong class anchors, centred per domain (first 200 1-shot episodes). Labels: accuracy on one unaugmented view per image with standard CLIP preprocessing; \cref{tab:desccomp} averages five views per image.}
  \label{fig:sim}
\end{figure}

\cref{fig:comp} in the main paper shows the distribution of frozen CLIP's support accuracy $c$: ISIC and ChestX episodes lie mostly below $\tau$, while EuroSAT and CropDiseases episodes lie mostly above it. The same gate operates independently on each episode's labeled support set, without a domain label or per-domain calibration. \cref{fig:sim} shows that frozen CLIP's similarities to correct versus competing class anchors are hard to distinguish on ChestX and ISIC, but separate more clearly on CropDiseases and EuroSAT. \cref{tab:desccomp} reports measured competence and gate opening rate for two anchor choices; these figures characterize the gate signal, not a domain-specific decision rule.

\begin{table}[H]
\centering
\caption{Competence estimates under two anchor constructions: frozen CLIP's mean support accuracy $c$ and the gate's open rate with class names versus description anchors (200 episodes at 1 shot and 400 at 5 shots).}
\label{tab:desccomp}
\small
\setlength{\tabcolsep}{5pt}
\begin{tabular}{llcccccccc}
\toprule
\rowcolor{headbg}
 & & \multicolumn{2}{c}{ISIC} & \multicolumn{2}{c}{EuroSAT} & \multicolumn{2}{c}{CropD} & \multicolumn{2}{c}{ChestX} \\
\rowcolor{headbg}
Shot & Anchors & $c$ & open & $c$ & open & $c$ & open & $c$ & open \\
\midrule
\multirow{2}{*}{1} & class names & 0.27 & 95\% & 0.51 & 34\% & 0.40 & 68\% & 0.20 & 100\% \\
\rowcolor{oursbg} & descriptions & 0.29 & 94\% & 0.63 & 12\% & 0.62 & 13\% & 0.20 & 100\% \\
\midrule
\multirow{2}{*}{5} & class names & 0.27 & 100\% & 0.52 & 26\% & 0.40 & 74\% & 0.21 & 100\% \\
\rowcolor{oursbg} & descriptions & 0.29 & 100\% & 0.63 & 8\% & 0.62 & 12\% & 0.20 & 100\% \\
\bottomrule
\end{tabular}
\end{table}

\FloatBarrier
\subsection{Trajectory Ensemble}

\cref{tab:traj} reads \ours{} out at epochs 30, 60 and 100 alone and with the ensemble. The ensemble gives the best average at both shots, and it removes the choice of a stopping epoch: the best single epoch differs across domains (epoch 30 on ISIC and epoch 60 on CropDiseases at 1 shot), while the ensemble stays within 0.6 points of the best single epoch on every domain.

\begin{table}[H]
\centering
\caption{Accuracy (\%) of \ours{} read out at one epoch and with the trajectory ensemble (all reported episodes).}
\label{tab:traj}
\small
\setlength{\tabcolsep}{4pt}
\begin{tabular}{llccccc}
\toprule
\rowcolor{headbg}
Shot & Readout & ISIC & EuroSAT & CropD & ChestX & Avg. \\
\midrule
\multirow{4}{*}{1} & epoch 30 & \textbf{43.02} & 85.15 & 87.40 & 22.58 & 59.54 \\
 & epoch 60 & 42.28 & 84.81 & 87.56 & 22.66 & 59.33 \\
 & epoch 100 & 42.01 & 84.39 & 87.52 & 22.66 & 59.14 \\
\rowcolor{oursbg} & ensemble & 42.49 & \textbf{85.29} & \textbf{87.78} & \textbf{22.67} & \textbf{59.56} \\
\midrule
\multirow{4}{*}{5} & epoch 30 & \textbf{58.02} & \textbf{93.67} & 97.20 & 26.06 & 68.74 \\
 & epoch 60 & 57.74 & 93.34 & \textbf{97.55} & 26.63 & 68.81 \\
 & epoch 100 & 57.66 & 92.99 & 97.41 & 26.58 & 68.66 \\
\rowcolor{oursbg} & ensemble & 58.01 & 93.56 & 97.42 & \textbf{26.86} & \textbf{68.96} \\
\bottomrule
\end{tabular}
\end{table}

\textbf{Shared ensemble control.} We reuse the matched ATHA and TIR runs in \cref{tab:main}, with 800 1-shot and 400 5-shot episodes per domain. We apply the same per-query, classwise z-score normalization as \ours{} to their logits saved at epochs 30, 60 and 100 and sum them, without retraining. At 1 and 5 shots, the BSCD-FSL mean accuracies are 55.70\% and 67.22\% for ATHA, and 56.99\% and 68.02\% for TIR.

\FloatBarrier
\subsection{Sensitivity}

\begin{figure}[H]
  \centering
  \includegraphics[width=\textwidth]{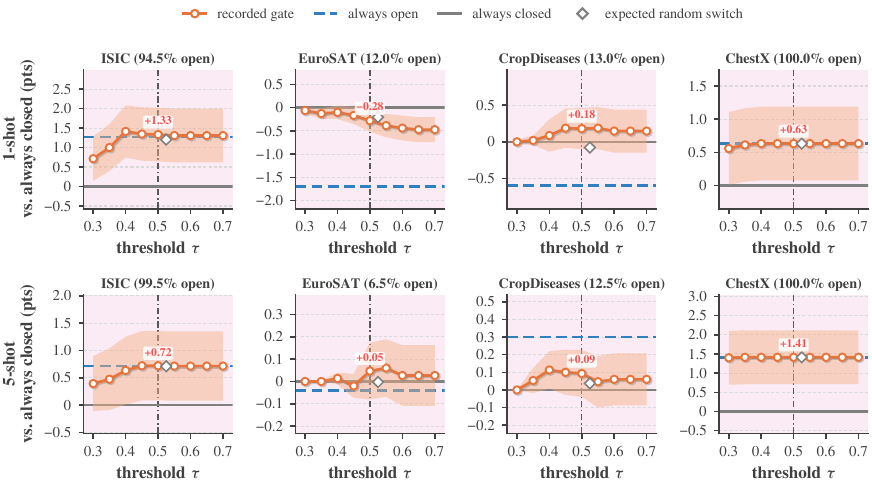}
    \caption{Gate minus always closing versus $\tau$ (first 200 episodes per domain; band: paired 95\% confidence interval). Dashed: always opening; diamond: expected random at the gate's open rate; titles: open rate at $\tau=0.5$. \cref{tab:desccomp} uses 400 episodes at 5 shots.}
  \label{fig:tau}
\end{figure}

\cref{tab:sens} varies every hyperparameter one at a time, and \cref{fig:tau} varies the threshold of the gate. All other parameters retain their default values.

\begin{table}[H]
\centering
\caption{Sensitivity to the hyperparameters (average accuracy, \%, first 100 episodes per domain; 5-shot unless noted; default underlined). $\beta$ and $\rho$ act only where the gate opens and are averaged over ISIC and ChestX, the fusion weight acts only where it closes and is averaged over EuroSAT and CropDiseases, and the other rows, including the open branch's learning-rate multiplier, are averaged over all four domains (on ChestX alone, the multiplier gives 26.33 at $\times$1 and 27.63 at $\times$2); $\tau$: Figure~\ref{fig:tau}.}
\label{tab:sens}
\footnotesize
\setlength{\tabcolsep}{3.2pt}
\begin{tabular}{lccccc}
\toprule
\rowcolor{headbg}LoRA lr multiplier (open) & $\times$1 & $\times$1.5 & $\times$2 & $\times$3 &  \\
Avg. & 69.17 & 69.23 & \underline{69.40} & 69.35 &  \\
\rowcolor{headbg}Crop min.\ scale & 0.3 & 0.5 & 0.8 &  &  \\
Avg. & 68.95 & \underline{69.40} & 69.40 &  &  \\
\rowcolor{headbg}Descriptions $M$ & 1 & 3 & 5 &  &  \\
Avg. & 69.29 & \underline{69.40} & 68.95 &  &  \\
\rowcolor{headbg}Fusion weight $\lambda$ (closed) & 0 & 0.25 & 0.5 & 0.75 & 1 \\
Avg. & 95.62 & 95.74 & \underline{95.47} & 79.96 & 68.88 \\
\rowcolor{headbg}Fusion weight $\lambda$, 1-shot & 0 & 0.25 & 0.5 &  &  \\
Avg. & 86.63 & 87.03 & \underline{86.93} &  &  \\
\rowcolor{headbg}Reweighting $\beta$ (open) & 0.25 & 0.5 & 0.75 &  &  \\
Avg. & 42.61 & \underline{43.33} & 43.01 &  &  \\
\rowcolor{headbg}Top-$k$ ratio $\rho$ (open) & 0.2 & 0.3 & 0.4 &  &  \\
Avg. & 42.67 & \underline{43.33} & 42.85 &  &  \\
\rowcolor{headbg}Reweighting $\beta$, 1-shot & 0.25 & 0.5 & 0.75 &  &  \\
Avg. & 33.21 & \underline{33.33} & 33.48 &  &  \\
\rowcolor{headbg}Top-$k$ ratio $\rho$, 1-shot & 0.2 & 0.3 & 0.4 &  &  \\
Avg. & 33.39 & \underline{33.33} & 33.03 &  &  \\
\bottomrule
\end{tabular}
\end{table}

\FloatBarrier
\section{Qualitative Examples}
\label{app:examples}

\begin{figure}[H]
  \centering
  \includegraphics[width=0.85\textwidth]{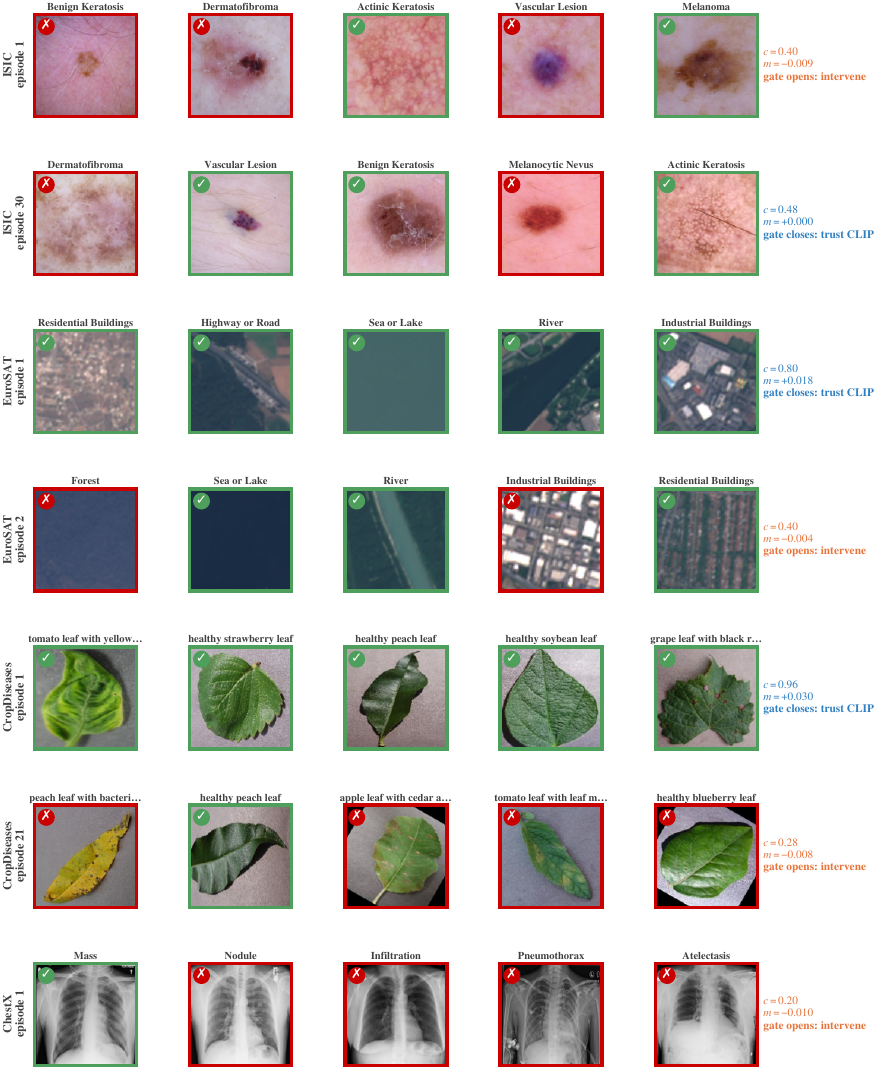}
  \caption{The gate on real 1-shot episodes, one per row. Marks: frozen CLIP's prediction on the plain image (\ding{51}: correct); $c$ and $m$ are computed on the five views of each image (\cref{eq:gate}).}
  \label{fig:exgate}
\end{figure}

\textbf{The gate on real episodes.} \cref{fig:exgate} shows the gate's decision on 1-shot evaluation episodes. For each BSCD-FSL domain, we take the first episode on which the gate takes the domain's majority decision and the first on which it takes the other one (ChestX has none: the gate opens on all its episodes). Frozen CLIP misreads most support images of ISIC and ChestX, and the gate opens; it reads most of EuroSAT and CropDiseases, and the gate closes. The minority cases follow the same rule: an ISIC episode whose support images frozen CLIP reads with a non-negative margin is left to CLIP, and EuroSAT and CropDiseases episodes that frozen CLIP misreads are intervened on.

\begin{figure}[H]
  \centering
  \includegraphics[width=0.95\textwidth]{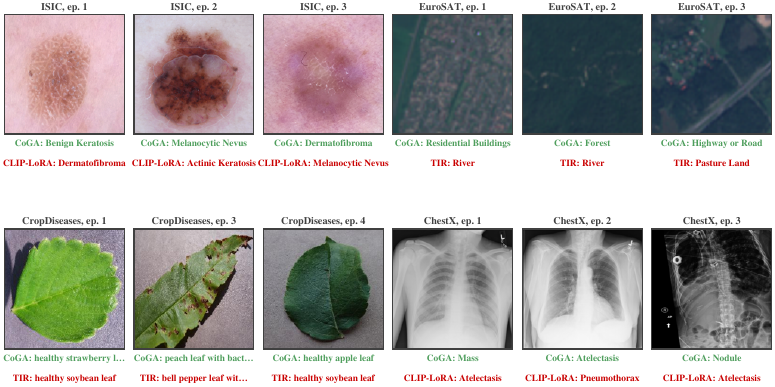}
  \caption{Queries corrected by \ours{} on the same episodes (green: \ours{}, the true class; red: the baseline, CLIP-LoRA where the gate opens and TIR where it closes).}
  \label{fig:expred}
\end{figure}

\FloatBarrier
\textbf{Corrected predictions.} \cref{fig:expred} shows query images that \ours{} classifies correctly and the baseline does not, on the same 1-shot episodes. Where the gate opens (ISIC, ChestX), the baseline is CLIP-LoRA, which the open branch improves; where it closes (EuroSAT, CropDiseases), the baseline is TIR, whose intervention the closed branch avoids. For each domain we take, in episode order, the first such query of each true class until three classes are shown. On EuroSAT and CropDiseases, TIR moves such queries to related classes, for example a residential area to river and a healthy strawberry leaf to healthy soybean leaf, and the closed branch keeps them in their class.

\FloatBarrier
\section{Class Descriptions}
\label{app:desc}

Each class anchor averages the text features of three prompts ``\{class name\}, \{description\}''; the descriptions were manually prepared, and folder names are replaced by the readable names below. The four BSCD-FSL domains come first, followed by the five further domains of \cref{tab:beyond} (DTD, FGVC-Aircraft, APTOS, Kvasir and PlantDoc).

{\small
\paragraph{ISIC2018.}
\begin{itemize}\setlength{\itemsep}{0pt}\setlength{\parskip}{0pt}
  \item \textit{Melanoma}: a dermoscopy image of an asymmetric lesion with irregular borders and several colors of brown, black, blue and red; an irregular dark pigmented skin lesion with uneven colors under dermoscopy; a dermoscopy image showing an atypical network, blue-white veil and irregular dots.
  \item \textit{Melanocytic Nevus}: a dermoscopy image of a symmetric round mole with uniform brown color and a regular pigment network; a regular round brown mole under dermoscopy; a dermoscopy image of a benign mole with an even reticular pattern.
  \item \textit{Basal Cell Carcinoma}: a dermoscopy image of a pinkish shiny lesion with branching blood vessels and blue-gray ovoid nests; a pearly pink lesion with arborizing vessels under dermoscopy; a dermoscopy image with blue-gray globules, ulceration and red branching vessels.
  \item \textit{Actinic Keratosis}: a dermoscopy image of a rough red scaly patch with a strawberry-like red pattern; a pink scaly sun-damaged skin patch under dermoscopy; a dermoscopy image of a red rough lesion with white scales around follicles.
  \item \textit{Benign Keratosis}: a dermoscopy image of a waxy brown stuck-on lesion with milia-like cysts and comedo-like openings; a light to dark brown warty lesion with a stuck-on look under dermoscopy; a dermoscopy image of a seborrheic keratosis with fissures and cysts.
  \item \textit{Dermatofibroma}: a dermoscopy image of a small firm lesion with a central white scar-like patch and a faint brown rim; a firm brownish nodule with a white center under dermoscopy; a dermoscopy image of a lesion with a central white patch and a delicate pigment ring.
  \item \textit{Vascular Lesion}: a dermoscopy image of a red to purple lesion made of well-defined red lagoons; a bright red or purple blood vessel lesion under dermoscopy; a dermoscopy image of an angioma with red-blue lacunae.
\end{itemize}
\paragraph{EuroSAT.}
\begin{itemize}\setlength{\itemsep}{0pt}\setlength{\parskip}{0pt}
  \item \textit{Annual Crop Land}: a satellite view of rectangular agricultural fields in brown and green stripes; a patchwork of plowed and planted fields seen from above; farmland with parallel tilled strips in a satellite image.
  \item \textit{Forest}: a satellite view of dense dark green tree canopy with a rough texture; a dark green woodland seen from above; a continuous mass of trees in a satellite image.
  \item \textit{Herbaceous Vegetation Land}: a satellite view of patchy light green and brown natural grassland and shrubs; mixed scrub and wild grassland seen from above; irregular natural vegetation with mottled green and tan tones.
  \item \textit{Highway or Road}: a satellite view of a long straight gray road crossing the land; a thin straight paved road through fields seen from above; a highway line crossing the landscape in a satellite image.
  \item \textit{Industrial Buildings}: a satellite view of large gray and white rectangular warehouse roofs and parking lots; big flat factory roofs and industrial lots seen from above; large bright rectangular buildings in a satellite image.
  \item \textit{Pasture Land}: a satellite view of a smooth uniform light green meadow; an even green grazing field seen from above; smooth grass fields with few features in a satellite image.
  \item \textit{Permanent Crop Land}: a satellite view of orchards or vineyards planted in regular rows of small dots; rows of fruit trees or vines seen from above; a regular grid of plants in an orchard in a satellite image.
  \item \textit{Residential Buildings}: a satellite view of a dense grid of small houses with red and gray roofs and streets; a neighborhood of houses and streets seen from above; dense small buildings and roads in a satellite image.
  \item \textit{River}: a satellite view of a winding dark blue or green river crossing the land; a narrow ribbon of water meandering through land seen from above; a curving river channel in a satellite image.
  \item \textit{Sea or Lake}: a satellite view of a large uniform dark blue water surface; open water filling the scene seen from above; a calm dark water body in a satellite image.
\end{itemize}
\paragraph{CropDiseases.}
\begin{itemize}\setlength{\itemsep}{0pt}\setlength{\parskip}{0pt}
  \item \textit{apple leaf with apple scab}: olive-green to dark brown velvety spots with feathery edges on the leaf; dark scabby lesions that make the apple leaf look rough and blotched; sooty olive spots spreading along the veins of the apple leaf.
  \item \textit{apple leaf with black rot}: circular frog-eye spots with purple edges and tan to brown centers; round brown lesions with a purple ring on the apple leaf; small purple specks that grow into tan spots with dark margins.
  \item \textit{apple leaf with cedar apple rust}: bright yellow-orange spots with red borders and tiny dark dots in the center; orange rust spots with a yellow halo on the apple leaf; vivid yellow and orange circular blotches on the upper surface.
  \item \textit{healthy apple leaf}: a smooth uniformly green leaf with no spots or lesions; a clean even green oval apple leaf with fine serrated edges; a fresh green apple leaf without discoloration.
  \item \textit{healthy blueberry leaf}: a small oval smooth uniformly green leaf with no spots; a clean even green elliptical leaf with a smooth margin; a fresh green blueberry leaf without discoloration.
  \item \textit{cherry leaf with powdery mildew}: white powdery fungal patches on the leaf surface, the leaf may curl; a cherry leaf dusted with white flour-like fungal growth; pale white fuzzy coating and slightly curled cherry leaf.
  \item \textit{healthy cherry leaf}: a glossy uniformly green leaf with serrated edges and no spots; a clean even green oval cherry leaf with fine teeth; a fresh green cherry leaf without discoloration.
  \item \textit{corn leaf with gray leaf spot}: long narrow rectangular gray to tan lesions running parallel to the veins; matchstick-like gray rectangular streaks on the corn leaf; pale gray blocky lesions confined between leaf veins.
  \item \textit{corn leaf with common rust}: many small reddish-brown powdery pustules scattered over the leaf; cinnamon-brown raised rust pustules dotting the corn leaf; scattered rusty red-brown specks on a green corn leaf.
  \item \textit{corn leaf with northern leaf blight}: long elliptical cigar-shaped gray-green to tan lesions on the leaf blade; large spindle-shaped grayish lesions several centimeters long on the corn leaf; long tan boat-shaped dead patches on the corn leaf.
  \item \textit{healthy corn leaf}: a long uniformly green leaf blade with parallel veins and no lesions; a clean even green strap-like corn leaf; a fresh green corn leaf without lesions or streaks.
  \item \textit{grape leaf with black rot}: circular reddish-brown spots with dark borders and small black dots; tan circular leaf spots with a black border on the grape leaf; brown round lesions dotted with black fruiting bodies.
  \item \textit{grape leaf with esca black measles}: tiger-stripe bands of yellow and reddish-brown between the veins; striped yellow and brown scorching between the veins of the grape leaf; interveinal browning that leaves green bands along the veins.
  \item \textit{grape leaf with leaf blight}: irregular dark brown to black angular spots that merge and dry the leaf; many dark brown irregular blotches covering the grape leaf; dry necrotic brown patches spreading over the grape leaf.
  \item \textit{healthy grape leaf}: a lobed uniformly green leaf with no spots; a clean even green grape leaf with lobes and toothed edges; a fresh green vine leaf without spots.
  \item \textit{orange leaf with citrus greening}: blotchy asymmetric mottled yellowing with yellow and green patches; yellow-green blotchy mottling on the citrus leaf; uneven yellowing with green veins on the orange leaf.
  \item \textit{peach leaf with bacterial spot}: small angular purple to brown spots, some with holes, on a narrow leaf; tiny dark angular spots and shot holes on the peach leaf; purple-brown speckles concentrated near the leaf tip.
  \item \textit{healthy peach leaf}: a long narrow lance-shaped uniformly green leaf; a clean even green slender pointed peach leaf; a fresh green peach leaf without spots.
  \item \textit{bell pepper leaf with bacterial spot}: small dark brown water-soaked spots with yellow halos; brown scabby spots with pale edges on the pepper leaf; small raised dark lesions scattered on the pepper leaf.
  \item \textit{healthy bell pepper leaf}: a smooth glossy uniformly green leaf; a clean even green oval pepper leaf; a fresh green pepper leaf without spots.
  \item \textit{potato leaf with early blight}: dark brown spots with concentric target-like rings and yellow halos; bullseye brown rings on the potato leaf; dark angular spots with concentric rings and yellowing tissue.
  \item \textit{potato leaf with late blight}: large irregular dark brown to black water-soaked lesions, often at the leaf edges; greasy brown-black blotches spreading from the potato leaf edge; dark wet-looking dead tissue on the potato leaf.
  \item \textit{healthy potato leaf}: a uniformly green compound leaf with no lesions; a clean even green potato leaflet; a fresh green potato leaf without lesions.
  \item \textit{healthy raspberry leaf}: a serrated uniformly green leaflet with no spots; a clean even green toothed raspberry leaf; a fresh green raspberry leaf without spots.
  \item \textit{healthy soybean leaf}: an oval uniformly green trifoliate leaf with no spots; a clean even green oval soybean leaflet; a fresh green soybean leaf without spots.
  \item \textit{squash leaf with powdery mildew}: white powdery patches covering a large broad leaf; a broad squash leaf coated with white powder; white dusty spots merging across the squash leaf.
  \item \textit{strawberry leaf with leaf scorch}: many small irregular dark purple spots that merge, the edges look burned; purple blotches and browned scorched edges on the strawberry leaf; dark purple speckles turning the strawberry leaf brown.
  \item \textit{healthy strawberry leaf}: a toothed uniformly green leaflet with no spots; a clean even green serrated strawberry leaflet; a fresh green strawberry leaf without spots.
  \item \textit{tomato leaf with bacterial spot}: many small dark brown to black greasy spots; tiny dark water-soaked specks peppered over the tomato leaf; small black spots with yellow halos on the tomato leaf.
  \item \textit{tomato leaf with early blight}: brown spots with concentric target-like rings surrounded by yellowing; bullseye brown rings on the lower tomato leaf; dark brown lesions with target rings and a yellow border.
  \item \textit{tomato leaf with late blight}: large irregular greasy gray-green to brown blotches; dark water-soaked lesions spreading over the tomato leaf; brown-black greasy patches with pale margins.
  \item \textit{tomato leaf with leaf mold}: pale yellow patches on top and olive-green velvety mold underneath; yellow blotches on the tomato leaf with fuzzy olive mold; pale yellow spots and velvety greenish-brown mold.
  \item \textit{tomato leaf with septoria leaf spot}: many small circular spots with gray centers and dark brown margins; numerous tiny round spots with dark rims and pale centers on the tomato leaf; small gray-centered speckles densely covering the tomato leaf.
  \item \textit{tomato leaf with spider mites}: fine yellow stippling and speckled bronze discoloration across the leaf; tiny yellow dots and faint webbing on a dull tomato leaf; bronzed speckled tomato leaf losing its green color.
  \item \textit{tomato leaf with target spot}: brown lesions with concentric rings and light centers; round brown target-like lesions with pale centers on the tomato leaf; brown spots with faint concentric zones on the tomato leaf.
  \item \textit{tomato leaf with yellow leaf curl virus}: small upward-curled crinkled leaves with yellow margins; yellowing curled-up tomato leaflets that look crumpled; small cupped tomato leaves with yellow edges.
  \item \textit{tomato leaf with mosaic virus}: a light and dark green mottled mosaic pattern on a distorted leaf; patchy light and dark green mottling on the tomato leaf; a puckered tomato leaf with a mosaic of green shades.
  \item \textit{healthy tomato leaf}: a uniformly green compound leaf with no spots; a clean even green tomato leaflet with toothed edges; a fresh green tomato leaf without spots.
\end{itemize}
\paragraph{ChestX.}
\begin{itemize}\setlength{\itemsep}{0pt}\setlength{\parskip}{0pt}
  \item \textit{Atelectasis}: a chest x-ray with a collapsed lung region of increased opacity and a shifted fissure or diaphragm; a frontal chest radiograph with a dense wedge of collapsed lung; a chest x-ray with volume loss and a white band of airless lung.
  \item \textit{Cardiomegaly}: a chest x-ray with an enlarged heart silhouette wider than half of the chest; a frontal chest radiograph with a large heart shadow; a chest x-ray where the heart takes up most of the chest width.
  \item \textit{Effusion}: a chest x-ray with fluid blunting the costophrenic angle and an opaque white lung base; a frontal chest radiograph with a white fluid layer at the lung bottom; a chest x-ray with a meniscus of fluid at the costophrenic angle.
  \item \textit{Infiltration}: a chest x-ray with hazy patchy white opacities spread over the lung fields; a frontal chest radiograph with ill-defined cloudy lung shadows; a chest x-ray with diffuse fluffy white patches in the lungs.
  \item \textit{Mass}: a chest x-ray with a large well-defined round white opacity in the lung; a frontal chest radiograph with a big round lung lesion; a chest x-ray with a large dense rounded shadow in the lung.
  \item \textit{Nodule}: a chest x-ray with a small round white spot in the lung; a frontal chest radiograph with a tiny round lung spot; a chest x-ray with a small well-circumscribed white dot.
  \item \textit{Pneumothorax}: a chest x-ray with a dark area without lung markings beside a visible lung edge; a frontal chest radiograph with air around a collapsed lung; a chest x-ray with a black region at the lung edge and no vessels.
\end{itemize}

\subsection*{Further domains}
{\footnotesize
\paragraph{DTD.}
\begin{itemize}\setlength{\itemsep}{0pt}\setlength{\parskip}{0pt}
  \item \textit{banded texture}: broad parallel bands of alternating color across the surface; a surface divided into wide uniform stripes; repeating bands of contrasting tones running in one direction.
  \item \textit{blotchy texture}: irregular patches of uneven color scattered over a surface; a surface covered with large shapeless blotches; mottled areas of light and dark with soft blurry edges.
  \item \textit{braided texture}: strands woven over and under each other in a plait; interlocking twisted strands forming a rope-like pattern; a chain of overlapping braided loops.
  \item \textit{bubbly texture}: many round bubbles of different sizes clustered together; a surface full of glossy spherical bubbles; foam-like clusters of transparent circles.
  \item \textit{bumpy texture}: a surface covered with small rounded raised bumps; an uneven surface of lumps and shallow mounds; a rough relief of irregular bulges casting small shadows.
  \item \textit{chequered texture}: alternating light and dark squares like a chessboard; a regular grid of two-colored square tiles; a checkerboard pattern of contrasting squares.
  \item \textit{cobwebbed texture}: thin silky threads of a spider web stretched across a gap; a radial web of fine strands with concentric rings; delicate white threads forming an irregular net.
  \item \textit{cracked texture}: a dry surface split by a network of jagged cracks; irregular polygons separated by dark fissures; a broken crust of cracked paint or mud.
  \item \textit{crosshatched texture}: two sets of fine parallel lines crossing at an angle; a dense pattern of intersecting diagonal strokes; criss-crossed hatching lines forming small diamonds.
  \item \textit{crystalline texture}: faceted crystals with sharp edges and glinting planes; a cluster of angular translucent crystal shapes; geometric mineral facets reflecting light.
  \item \textit{dotted texture}: small evenly spaced dots on a plain background; a pattern of tiny round spots in rows; a surface sprinkled with uniform small points.
  \item \textit{fibrous texture}: many thin threads or fibers running in loose strands; a tangle of fine hair-like filaments; a surface made of long stringy fibers.
  \item \textit{flecked texture}: tiny specks of contrasting color scattered over a surface; small irregular flecks across a uniform background; a speckled surface with fine spots of color.
  \item \textit{freckled texture}: small brown spots scattered over skin or a pale surface; many light brown freckles of varying size; a pale surface dotted with tan speckles.
  \item \textit{frilly texture}: ruffled wavy edges of fabric folded in layers; delicate gathered frills with curling edges; rows of soft ruffles like lace trim.
  \item \textit{gauzy texture}: thin sheer translucent fabric with a soft blur; a light see-through veil of fine cloth; delicate semi-transparent material with faint folds.
  \item \textit{grid texture}: straight lines crossing at right angles forming squares; a regular lattice of horizontal and vertical lines; a uniform square mesh of thin bars.
  \item \textit{grooved texture}: long parallel channels cut into a surface; a ridged surface of evenly spaced grooves; a series of narrow troughs and ridges side by side.
  \item \textit{honeycombed texture}: a regular array of hexagonal cells; a honeycomb of six-sided hollow compartments; tessellated hexagons with thin walls.
  \item \textit{interlaced texture}: strands crossing over and under each other in a weave; intertwined bands forming a knotted lattice; loops and strips threaded through one another.
  \item \textit{knitted texture}: rows of small interlocking V-shaped yarn loops; the stitches of a knitted wool fabric; soft textured knit loops in tight rows.
  \item \textit{lacelike texture}: an openwork pattern of delicate holes and threads like lace; fine ornamental lace with floral cutouts; a filigree of thin threads and small openings.
  \item \textit{lined texture}: many thin parallel lines across a surface; a surface ruled with straight narrow lines; repeated fine lines with even spacing.
  \item \textit{marbled texture}: swirling veins of color through polished stone; a smooth surface with flowing marble streaks; white and grey marble with irregular veining.
  \item \textit{matted texture}: a dense flattened tangle of hair or fibers; compressed felt-like strands pressed together; a messy mat of interwoven fibers.
  \item \textit{meshed texture}: a net of thin wires or threads with regular openings; an open mesh of small diamond or square holes; a fine netting pattern seen up close.
  \item \textit{paisley texture}: teardrop-shaped motifs with curved tips and ornate fill; a decorative paisley pattern of curling droplets; repeated swirling teardrop shapes on fabric.
  \item \textit{perforated texture}: a flat sheet pierced by many round holes in rows; a surface punched with evenly spaced holes; a metal plate with a regular pattern of perforations.
  \item \textit{pitted texture}: a surface marked with many small hollows and pits; a rough surface of shallow craters; a corroded surface dotted with depressions.
  \item \textit{pleated texture}: fabric folded into sharp regular pleats; parallel creased folds like an accordion; evenly spaced pressed folds in cloth.
  \item \textit{polka-dotted texture}: large round dots evenly spread on a plain background; a fabric with regularly spaced circular spots; bold polka dots in a repeating grid.
  \item \textit{porous texture}: a spongy surface full of tiny irregular holes; a material riddled with small pores; a sponge-like texture of open cavities.
  \item \textit{potholed texture}: a road surface broken by rough holes; worn asphalt with deep uneven potholes; a damaged pavement with water-filled cavities.
  \item \textit{scaly texture}: overlapping scales like the skin of a fish or reptile; rows of small plate-like scales; a surface tiled with curved overlapping scales.
  \item \textit{smeared texture}: streaks of paint or dirt wiped across a surface; blurred smudges with dragged edges; a surface with long smeared marks of color.
  \item \textit{spiralled texture}: a curve winding outward from a center in a spiral; coiled spiral shapes like a shell or spring; repeating spiral swirls.
  \item \textit{sprinkled texture}: small colorful bits scattered over a surface like sprinkles; tiny elongated specks of many colors; a surface topped with scattered sprinkles.
  \item \textit{stained texture}: irregular discolored marks soaked into a surface; blotches of stain with uneven darker rims; a surface spoiled by spreading stains.
  \item \textit{stratified texture}: horizontal layers of rock or sediment stacked in bands; layered strata of different colors and thicknesses; a cliff face of parallel sedimentary layers.
  \item \textit{striped texture}: regular stripes of alternating color; bold parallel stripes like a zebra or fabric; evenly spaced contrasting stripes.
  \item \textit{studded texture}: a surface set with raised metal studs or rivets; rows of small round knobs protruding from leather or metal; a pattern of evenly spaced studs.
  \item \textit{swirly texture}: flowing swirls and curls of color; soft twisting curves blending into each other; a pattern of curling wavy swirls.
  \item \textit{veined texture}: a network of branching veins like a leaf; thin branching lines spreading across a surface; a leaf with a visible network of veins.
  \item \textit{waffled texture}: a grid of square indentations like a waffle; raised ridges forming small square pockets; a waffle pattern of deep square cells.
  \item \textit{woven texture}: threads interlaced in a regular over-under weave; a basket or fabric weave of crossing strips; a tight woven pattern of warp and weft.
  \item \textit{wrinkled texture}: a crumpled surface with many irregular creases; folds and wrinkles in skin or fabric; a creased surface with uneven ridges.
  \item \textit{zigzagged texture}: sharp back-and-forth zigzag lines; a chevron pattern of angled stripes; repeated V-shaped zigzags across a surface.
\end{itemize}
\paragraph{FGVC-Aircraft.}
\begin{itemize}\setlength{\itemsep}{0pt}\setlength{\parskip}{0pt}
  \item \textit{Boeing 707-320}: an early four-engine jet airliner with engines in pods under swept wings; a narrow-body jet with a long slender fuselage and a tall fin with a forward antenna; a vintage four-jet airliner with a pointed nose.
  \item \textit{Boeing 727-200}: a trijet airliner with three engines at the rear and a T-tail; a narrow-body jet with an S-shaped center engine intake at the base of the fin; a stretched trijet with no engines under the wings.
  \item \textit{Boeing 737-200}: a short twin-jet airliner with slim cigar-shaped engines hugging the wing; an early 737 with long narrow engine nacelles under the wings; a small narrow-body jet with a low stance.
  \item \textit{Boeing 737-300}: a twin-jet airliner with flattened oval engine intakes under the wings; a classic 737 with a hamster-pouch engine nacelle and a fin fillet; a short narrow-body jet with two wing-mounted engines.
  \item \textit{Boeing 737-400}: a stretched classic 737 with flattened engine intakes; a narrow-body twin-jet slightly longer than the 737-300; a twin-engine airliner with oval nacelles and a long cabin window row.
  \item \textit{Boeing 737-500}: the shortest classic 737 with flattened oval engine intakes; a stubby narrow-body twin-jet; a compact twin-engine airliner with a short fuselage.
  \item \textit{Boeing 737-600}: a short next-generation 737 with round engine intakes; a compact twin-jet with a tall fin and no winglets; the smallest next-generation 737 with a short fuselage.
  \item \textit{Boeing 737-700}: a next-generation 737 of medium length, often with blended winglets; a twin-jet airliner with round engine intakes and upturned wingtips; a narrow-body twin-engine jet with a modest fuselage.
  \item \textit{Boeing 737-800}: a long next-generation 737 with blended winglets; a stretched twin-jet with two over-wing exits on each side; a narrow-body airliner with round engines and a long cabin.
  \item \textit{Boeing 737-900}: the longest next-generation 737 with winglets; a very long narrow-body twin-jet; a stretched twin-engine airliner with a long window row.
  \item \textit{Boeing 747-100}: a four-engine jumbo jet with a short upper-deck hump; a wide-body airliner with a humped forward fuselage and few upper windows; an early jumbo jet with four wing-mounted engines.
  \item \textit{Boeing 747-200}: a four-engine jumbo jet with a short upper-deck hump; a wide-body airliner with four large engines under the wings; a classic jumbo jet with a humped nose section.
  \item \textit{Boeing 747-300}: a jumbo jet with a stretched upper deck and a long hump; a four-engine wide-body with many upper-deck windows; a classic 747 with an extended upper deck.
  \item \textit{Boeing 747-400}: a four-engine jumbo jet with a long upper deck and raised winglets; a wide-body airliner with a hump and upturned wingtips; a large four-engine jet with a stretched upper deck.
  \item \textit{Boeing 757-200}: a long narrow-body twin-jet with a tall landing gear stance; a slender single-aisle airliner with two large engines and a pointed nose; a narrow-body jet with a long thin fuselage.
  \item \textit{Boeing 757-300}: an extremely long narrow-body twin-jet; a stretched single-aisle airliner with two large wing engines; a very long slender 757 fuselage.
  \item \textit{Boeing 767-200}: a short wide-body twin-jet with two large engines; a twin-aisle airliner with a stubby rounded fuselage; a medium wide-body jet with a tall fin.
  \item \textit{Boeing 767-300}: a wide-body twin-jet of medium length; a twin-engine twin-aisle airliner longer than the 767-200; a wide-body jet with two large turbofans under the wings.
  \item \textit{Boeing 767-400}: a long wide-body twin-jet with raked swept wingtips; the stretched 767 with angled wingtip extensions; a long twin-engine wide-body airliner.
  \item \textit{Boeing 777-200}: a large wide-body twin-jet with very big round engines; a twin-engine airliner with six-wheel main landing gear bogies; a wide-body with a blade-like tail cone.
  \item \textit{Boeing 777-300}: a very long wide-body twin-jet with huge engines; a stretched 777 with six-wheel main gear; a long twin-engine twin-aisle airliner.
  \item \textit{Airbus A300B4}: an early Airbus wide-body twin-jet with a rounded nose; a twin-engine wide-body with a plain wingtip and a stubby fuselage; a first-generation twin-aisle airliner.
  \item \textit{Airbus A310}: a short Airbus wide-body twin-jet, often with small wingtip fences; a shortened twin-aisle airliner with two engines; a compact wide-body jet with a rounded nose.
  \item \textit{Airbus A318}: the shortest Airbus narrow-body, a stubby twin-jet with a tall fin; a very short single-aisle airliner; a small twin-engine jet with small wingtip fences.
  \item \textit{Airbus A319}: a short Airbus narrow-body twin-jet; a single-aisle airliner with wingtip fences and a short cabin; a compact twin-engine jet with a rounded nose.
  \item \textit{Airbus A320}: an Airbus narrow-body twin-jet with wingtip fences or sharklets; a single-aisle airliner with two wing engines and a rounded nose; a medium-length twin-engine jet.
  \item \textit{Airbus A321}: a long Airbus narrow-body twin-jet; a stretched single-aisle airliner with extra exit doors; a long twin-engine jet with a rounded nose.
  \item \textit{Airbus A330-200}: a wide-body twin-jet with long wings and raised winglets; a twin-engine twin-aisle airliner with a shorter fuselage; a large Airbus with two engines and a rounded nose.
  \item \textit{Airbus A330-300}: a long wide-body twin-jet with winglets; a stretched twin-engine twin-aisle airliner; a large Airbus with two big engines.
  \item \textit{Airbus A340-200}: a four-engine wide-body with slim engines and winglets; a shorter four-engine Airbus long-haul airliner; a twin-aisle jet with four small engines under the wings.
  \item \textit{Airbus A340-300}: a four-engine wide-body with four slim engines; a long-haul Airbus with four engines and winglets; a twin-aisle airliner with a long fuselage and four engines.
  \item \textit{Airbus A340-500}: a long four-engine Airbus with larger engines; a stretched four-engine wide-body jet; a long-range airliner with four big turbofans.
  \item \textit{Airbus A340-600}: a very long four-engine Airbus wide-body; an extremely long four-engine airliner; a stretched twin-aisle jet with four large engines.
  \item \textit{Airbus A380}: a huge double-deck airliner with two full rows of windows; a four-engine superjumbo with a deep fuselage; a massive wide-body with full-length upper deck.
  \item \textit{ATR 42}: a small high-wing twin turboprop with a T-tail; a short regional propeller airliner; a twin-propeller plane with wings on top of the fuselage.
  \item \textit{ATR 72}: a stretched high-wing twin turboprop with a T-tail; a long regional propeller airliner; a twin-propeller plane with a high wing and long cabin.
  \item \textit{Antonov An-12}: a four-engine turboprop military transport with a high wing; a cargo plane with a glazed nose and a tail gun position; a high-wing four-propeller freighter.
  \item \textit{BAe 146-200}: a four-engine high-wing regional jet with a T-tail; a small airliner with four engines under a high wing; a stubby high-wing jet with a T-tail.
  \item \textit{BAe 146-300}: a stretched four-engine high-wing regional jet; a high-wing jet with four engines and a T-tail; a long-bodied 146 regional airliner.
  \item \textit{BAe 125}: a small business jet with two rear-mounted engines; a corporate jet with a low swept wing and rear engines; a compact twin-jet with a mid T-tail.
  \item \textit{Beechcraft 1900}: a small twin turboprop commuter airliner; a slender propeller plane with a T-tail and small fins; a narrow fuselage twin-prop regional aircraft.
  \item \textit{Boeing 717}: a small twin-jet with two engines at the rear and a T-tail; a short airliner with rear-mounted engines; a compact T-tail jet with clean wings.
  \item \textit{Lockheed C-130 Hercules}: a four-engine turboprop military transport with a high wing; a stout grey cargo plane with an upswept tail; a high-wing four-propeller freighter.
  \item \textit{Douglas C-47}: a vintage twin-propeller transport with a tail-wheel; a World War II era cargo plane with round engines; an old low-wing twin-prop with a sloping stance on the ground.
  \item \textit{Bombardier CRJ-200}: a small regional jet with two rear engines and a T-tail; a slim fuselage regional airliner with rear-mounted engines; a short T-tail jet with winglets.
  \item \textit{Bombardier CRJ-700}: a stretched regional jet with rear engines and a T-tail; a slim regional airliner longer than the CRJ-200; a T-tail jet with two engines on the rear fuselage.
  \item \textit{Bombardier CRJ-900}: a long regional jet with rear engines and a T-tail; a slim stretched regional airliner; a long T-tail jet with two rear engines.
  \item \textit{Cessna 172}: a small single-engine high-wing propeller plane; a light four-seat plane with a strut-braced high wing; a small private propeller airplane with fixed landing gear.
  \item \textit{Cessna 208 Caravan}: a single-engine turboprop with a high strut-braced wing; a utility plane with a long nose and boxy cabin; a high-wing propeller plane sometimes on floats.
  \item \textit{Cessna 525 CitationJet}: a light business jet with two rear engines and a straight wing; a small corporate jet with a T-shaped tail; a compact twin-jet with a short fuselage.
  \item \textit{Cessna Citation 560}: a business jet with two rear engines and a straight wing; a mid-size corporate twin-jet; a small jet with a cruciform tail.
  \item \textit{Bombardier Challenger 600}: a wide-cabin business jet with two rear engines and a T-tail; a corporate jet with winglets and a T-tail; a business twin-jet with a stout fuselage.
  \item \textit{McDonnell Douglas DC-10}: a wide-body trijet with one engine through the base of the tail fin; a three-engine airliner with two engines under the wings and one in the tail; a large trijet with a straight-through tail engine.
  \item \textit{Douglas DC-3}: a vintage twin-propeller airliner with a tail-wheel; a classic low-wing twin-prop with a rounded nose; an old propeller airliner sitting nose-high on the ground.
  \item \textit{Douglas DC-6}: a four-engine piston propeller airliner; a vintage airliner with four radial engines and a tricycle gear; a classic four-prop transport with a long fuselage.
  \item \textit{Douglas DC-8}: an early four-engine jet airliner with podded engines; a long narrow-body four-jet; a vintage jet with four engines under swept wings.
  \item \textit{Douglas DC-9-30}: a narrow-body twin-jet with rear engines and a T-tail; a slim airliner with two engines on the rear fuselage; a classic T-tail jet with clean wings.
  \item \textit{de Havilland Tiger Moth}: a small vintage biplane with two stacked wings; an open-cockpit biplane trainer; an old fabric-covered biplane with a tail-wheel.
  \item \textit{de Havilland Canada Chipmunk}: a small single-engine tandem-seat trainer with a tail-wheel; a low-wing piston propeller trainer with a canopy; a vintage light trainer plane.
  \item \textit{de Havilland Canada Twin Otter}: a high-wing twin turboprop utility plane with fixed gear; a boxy short-takeoff propeller plane, sometimes on floats; a strut-braced high-wing twin-prop.
  \item \textit{de Havilland Canada Dash 8-100}: a short high-wing twin turboprop with a T-tail; a regional propeller airliner with engines under a high wing; a compact Dash 8 turboprop.
  \item \textit{de Havilland Canada Dash 8-300}: a stretched high-wing twin turboprop with a T-tail; a longer regional propeller airliner; a Dash 8 with a long cabin and high wings.
  \item \textit{Robin DR400}: a small single-engine low-wing plane with cranked wingtips; a light touring propeller aircraft with a bubble canopy; a small wooden propeller plane with upturned outer wings.
  \item \textit{Dornier 328}: a high-wing twin turboprop regional plane with a T-tail; a small regional propeller airliner with a slim fuselage; a high-wing commuter turboprop.
  \item \textit{Embraer E170}: a regional jet with two engines under the wings; a small single-aisle airliner with a double-bubble fuselage; a compact twin-jet regional airliner.
  \item \textit{Embraer E190}: a stretched regional jet with two wing-mounted engines; a single-aisle airliner larger than the E170; a twin-jet with a long slim fuselage.
  \item \textit{Embraer E195}: the longest E-jet with two wing-mounted engines; a stretched regional airliner; a long single-aisle twin-jet.
  \item \textit{Embraer EMB 120 Brasilia}: a small twin turboprop with a T-tail and low wing; a regional propeller plane with a slim fuselage; a commuter turboprop with a T-tail.
  \item \textit{Embraer ERJ 135}: a short regional jet with two rear engines and a T-tail; a slim fuselage jet with engines on the rear; a small T-tail regional jet.
  \item \textit{Embraer ERJ 145}: a slender regional jet with two rear engines and a T-tail; a long thin fuselage with rear-mounted engines; a T-tail regional jet with a pointed nose.
  \item \textit{Embraer Legacy 600}: a business jet with rear engines, a T-tail and winglets; a corporate jet derived from a regional airliner; a long slim private jet with rear engines.
  \item \textit{Eurofighter Typhoon}: a delta-wing fighter jet with canard foresurfaces; a twin-engine combat jet with a chin air intake; a single-fin delta fighter with small front wings.
  \item \textit{F-16 Fighting Falcon}: a single-engine fighter jet with a belly air intake; a fighter with a bubble canopy and a single vertical fin; a small agile combat jet with cropped delta wings.
  \item \textit{F/A-18 Hornet}: a twin-engine fighter jet with two canted vertical tails; a carrier fighter with long leading-edge extensions; a grey combat jet with twin fins.
  \item \textit{Dassault Falcon 2000}: a business jet with two rear engines; a wide-cabin corporate twin-jet; a private jet with a cruciform tail.
  \item \textit{Dassault Falcon 900}: a business jet with three rear engines, one in the tail; a corporate trijet with an S-duct center engine; a private jet with three engines at the back.
  \item \textit{Fokker 100}: a narrow-body twin-jet with rear engines and a T-tail; a regional airliner with engines on the rear fuselage; a slim T-tail jet.
  \item \textit{Fokker 50}: a high-wing twin turboprop regional airliner; a propeller plane with engines under a high wing and a conventional tail; a regional turboprop with six-bladed propellers.
  \item \textit{Fokker 70}: a short twin-jet with rear engines and a T-tail; a compact regional airliner with rear-mounted engines; a shortened Fokker 100.
  \item \textit{Bombardier Global Express}: a large long-range business jet with rear engines and a T-tail; a corporate jet with long swept wings and winglets; a big private twin-jet.
  \item \textit{Gulfstream IV}: a large business jet with oval cabin windows and rear engines; a corporate twin-jet with a T-tail and winglets; a long-range private jet.
  \item \textit{Gulfstream V}: a long-range business jet with oval windows and rear engines; a stretched Gulfstream with a T-tail; a large corporate twin-jet.
  \item \textit{BAE Hawk T1}: a small jet trainer with a tandem canopy; a red or black military trainer jet with low wings; a single-engine trainer with side intakes.
  \item \textit{Ilyushin Il-76}: a four-engine military jet transport with a high wing and T-tail; a cargo jet with a glazed navigator nose; a heavy freighter with drooping wings and four engines.
  \item \textit{Lockheed L-1011 TriStar}: a wide-body trijet with an S-duct engine in the tail; a three-engine airliner with two wing engines and one at the rear; a large trijet with a center engine intake above the fuselage.
  \item \textit{McDonnell Douglas MD-11}: a wide-body trijet with winglets and a tail-fin engine; a stretched DC-10 with upturned wingtips; a three-engine long-haul airliner.
  \item \textit{McDonnell Douglas MD-80}: a long narrow-body twin-jet with rear engines and a T-tail; a slim airliner with long engine pods at the rear; a stretched DC-9 style jet.
  \item \textit{McDonnell Douglas MD-87}: a shortened MD-80 with rear engines and a T-tail; a compact narrow-body twin-jet; a T-tail airliner with a short fuselage.
  \item \textit{McDonnell Douglas MD-90}: a long narrow-body twin-jet with large rear engines and a T-tail; an MD-80 style airliner with bigger engines; a slim T-tail jet.
  \item \textit{Fairchild Metroliner}: a small twin turboprop with a long thin fuselage; a commuter propeller plane with a pointed nose; a narrow tube-like turboprop.
  \item \textit{Beechcraft King Air B200}: a twin turboprop business plane with a T-tail; a small propeller aircraft with a low wing; a corporate twin-prop.
  \item \textit{Piper PA-28 Cherokee}: a small single-engine low-wing propeller plane; a light four-seat private plane; a single-prop trainer with low wings.
  \item \textit{Cirrus SR20}: a small single-engine low-wing composite plane; a light propeller aircraft with a sleek fuselage; a modern four-seat single-prop.
  \item \textit{Saab 2000}: a twin turboprop regional airliner with a low wing; a long slim propeller airliner with six-bladed propellers; a fast regional turboprop.
  \item \textit{Saab 340}: a small twin turboprop regional airliner with a low wing; a commuter propeller plane; a compact regional turboprop.
  \item \textit{Supermarine Spitfire}: a World War II fighter with elliptical wings; a single-engine propeller fighter with a tail-wheel; a vintage camouflaged fighter plane.
  \item \textit{Panavia Tornado}: a variable-sweep wing combat jet; a twin-engine strike aircraft with a tall fin; a military jet with swing wings.
  \item \textit{Tupolev Tu-134}: a twin-jet airliner with rear engines and a T-tail and a glazed nose; a Soviet airliner with swept wings and rear engines; a narrow-body jet with a T-tail.
  \item \textit{Tupolev Tu-154}: a trijet airliner with three rear engines and a T-tail; a Soviet narrow-body jet with landing gear pods on the wings; a three-engine rear-engined airliner.
  \item \textit{Yakovlev Yak-42}: a trijet regional airliner with rear engines and a T-tail; a Soviet three-engine narrow-body; a regional jet with three engines at the tail.
\end{itemize}
\paragraph{APTOS.}
\begin{itemize}\setlength{\itemsep}{0pt}\setlength{\parskip}{0pt}
  \item \textit{fundus photo with mild diabetic retinopathy}: a retinal fundus photo with a few tiny red dots of microaneurysms; an orange retina with sparse small red spots near the vessels; a mostly normal fundus with only scattered pinpoint red dots.
  \item \textit{fundus photo with moderate diabetic retinopathy}: a retinal fundus photo with several red blot hemorrhages and yellow exudates; an orange retina with scattered red spots and small yellow patches; a fundus with dot hemorrhages and hard exudates around the macula.
  \item \textit{fundus photo of a healthy retina}: a clear orange-red fundus with a bright optic disc and smooth vessels; a uniform retina without red spots or yellow deposits; a healthy fundus photo with a sharp optic disc and clean blood vessels.
  \item \textit{fundus photo with proliferative diabetic retinopathy}: a fundus photo with abnormal new blood vessels and large hemorrhages; a retina with fibrous white tissue and tangled new vessels near the optic disc; a fundus with dark red bleeding patches and irregular vessel growth.
  \item \textit{fundus photo with severe diabetic retinopathy}: a fundus photo with many red hemorrhages in all quadrants; a retina with beaded veins and numerous blot hemorrhages; a fundus covered with dense red spots and white cotton-wool patches.
\end{itemize}
\paragraph{Kvasir.}
\begin{itemize}\setlength{\itemsep}{0pt}\setlength{\parskip}{0pt}
  \item \textit{endoscopy of Barrett's esophagus}: salmon-pink tongues of mucosa extending up the esophagus; reddish velvety patches above the gastroesophageal junction; irregular pink-red mucosa replacing the pale esophageal lining.
  \item \textit{endoscopy of short-segment Barrett's esophagus}: a short tongue of salmon-colored mucosa just above the z-line; a small reddish extension at the esophagogastric junction; a brief irregular pink band above the junction of esophagus and stomach.
  \item \textit{poorly prepared colon}: a colon view obscured by brown stool and murky fluid; a bowel lumen filled with solid feces hiding the mucosa; a dirty colon with large amounts of residual stool.
  \item \textit{well prepared colon}: a clean pink colon wall with clearly visible folds; a clear bowel lumen with only small traces of fluid; a well cleaned colon with visible mucosa and vessels.
  \item \textit{endoscopy of the cecum}: the cecum with the appendiceal orifice and converging folds; a pale pink pouch with a crescent-shaped appendix opening; the end of the colon with the fold of the ileocecal valve.
  \item \textit{dyed lifted polyp}: a polyp raised by a blue-dyed injection cushion; a bulging mucosa stained bright blue under a polyp; a lifted lesion on a light blue background before removal.
  \item \textit{dyed resection margin}: a blue-stained wound bed after polyp removal; a flat defect with blue dye at its edges; a resection site with blue tinted margins and exposed submucosa.
  \item \textit{mild esophagitis}: short red streaks of erosion along the esophageal folds; small linear red breaks in the pale esophageal lining; a few thin red stripes near the gastroesophageal junction.
  \item \textit{severe esophagitis}: long confluent red erosions covering the lower esophagus; extensive inflamed red mucosa with whitish exudate; wide merging red ulcerated streaks around the esophagus.
  \item \textit{impacted stool in the colon}: a colon lumen blocked by a hard brown stool mass; dense dark fecal material filling the view; compacted brown stool pressing against the colon wall.
  \item \textit{colon polyp}: a rounded pink growth protruding from the colon wall; a raised mushroom-like lump on the intestinal mucosa; a smooth reddish bump sticking out into the bowel lumen.
  \item \textit{endoscopy of the pylorus}: a round opening at the bottom of the stomach surrounded by smooth folds; the pyloric canal as a dark central hole in pink mucosa; a ring-shaped sphincter leading out of the stomach.
  \item \textit{retroflexed view of the rectum}: the black endoscope shaft seen inside the rectum; the endoscope looking back at itself inside a pink rectum; a close view of the anal canal with the scope tube visible.
  \item \textit{retroflexed view of the stomach}: the endoscope shaft entering the upper stomach; the black scope tube seen from inside the gastric fundus; large gastric folds around the endoscope at the cardia.
  \item \textit{very mild ulcerative colitis}: colon mucosa with slight redness and faint loss of vessel pattern; a mildly inflamed pink bowel wall with few changes; a nearly normal colon lining with subtle redness.
  \item \textit{mild ulcerative colitis}: a mildly red and swollen colon lining with reduced vessel visibility; slightly granular reddened mucosa; a colon wall with mild redness and edema.
  \item \textit{moderate ulcerative colitis}: a red friable colon wall with erosions and no visible vessels; moderately inflamed mucosa with small ulcers; marked redness and granular texture of the colon lining.
  \item \textit{moderate to severe ulcerative colitis}: a strongly inflamed colon with erosions and patchy bleeding; red granular mucosa with scattered ulcers and exudate; an inflamed colon wall with bleeding spots.
  \item \textit{severe ulcerative colitis}: a severely ulcerated colon with spontaneous bleeding; large ulcers and dark red bleeding mucosa; a raw inflamed colon wall covered in blood and exudate.
  \item \textit{endoscopy of the z-line}: the wavy junction between pale esophageal and red gastric mucosa; a clear border where the white esophageal lining meets the pink stomach lining; an irregular serrated line at the esophagogastric junction.
\end{itemize}
\paragraph{PlantDoc.}
\begin{itemize}\setlength{\itemsep}{0pt}\setlength{\parskip}{0pt}
  \item \textit{apple leaf with apple scab}: olive-green to black velvety spots on an apple leaf; dark scabby blotches spreading across an apple leaf; an apple leaf with rough brown-black lesions.
  \item \textit{healthy apple leaf}: a healthy green apple leaf with a serrated edge; a smooth uniformly green oval apple leaf; a clean apple leaf with no spots.
  \item \textit{apple leaf with rust}: an apple leaf with bright orange-yellow round spots; rusty orange lesions with red borders on an apple leaf; an apple leaf dotted with yellow-orange rust pustules.
  \item \textit{healthy bell pepper leaf}: a healthy glossy green bell pepper leaf; a smooth elongated pepper leaf with even color; a clean dark green pepper leaf.
  \item \textit{bell pepper leaf with leaf spot}: a pepper leaf with small brown water-soaked spots; dark lesions with yellow halos on a pepper leaf; a bell pepper leaf speckled with brown spots.
  \item \textit{healthy blueberry leaf}: a small oval smooth green blueberry leaf; glossy elliptical blueberry leaves on a twig; a healthy blueberry leaf with a fine vein pattern.
  \item \textit{healthy cherry leaf}: a healthy glossy green cherry leaf with a serrated margin; an elongated pointed cherry leaf; a clean bright green cherry leaf.
  \item \textit{corn leaf with gray leaf spot}: a corn leaf with long rectangular gray-tan lesions between the veins; narrow grayish stripes along a maize leaf; a corn leaf with parallel pale gray blotches.
  \item \textit{corn leaf with leaf blight}: a corn leaf with long cigar-shaped tan lesions; large elongated gray-brown dead patches on a maize leaf; a blighted corn leaf with spreading tan streaks.
  \item \textit{corn leaf with rust}: a corn leaf covered with small reddish-brown pustules; rusty brown powdery dots scattered on a maize leaf; a corn leaf speckled with cinnamon-colored rust spots.
  \item \textit{healthy peach leaf}: a long narrow healthy green peach leaf; a lance-shaped glossy peach leaf; slender green peach leaves with fine serrations.
  \item \textit{potato leaf with early blight}: a potato leaf with brown spots showing concentric rings; target-like dark lesions with yellow halos on a potato leaf; a potato leaf with dry brown bullseye spots.
  \item \textit{potato leaf with late blight}: a potato leaf with large dark water-soaked patches; brown-black irregular lesions spreading over a potato leaf; a potato leaf with blackened decaying areas and pale edges.
  \item \textit{healthy raspberry leaf}: a green raspberry leaf with a wrinkled surface and toothed edges; a compound raspberry leaf with three leaflets; a healthy textured raspberry leaf.
  \item \textit{healthy soybean leaf}: a healthy green trifoliate soybean leaf; an oval soybean leaflet with smooth margins; clean uniform green soybean leaves.
  \item \textit{squash leaf with powdery mildew}: a squash leaf covered with white powdery patches; a large broad leaf dusted with a white fungal coating; a squash leaf with chalky white spots spreading over it.
  \item \textit{healthy strawberry leaf}: a healthy strawberry leaf with three toothed leaflets; a bright green serrated strawberry leaf; a clean trifoliate strawberry leaf.
  \item \textit{tomato leaf with early blight}: a tomato leaf with brown spots showing concentric rings; target-like dark lesions with yellowing on a tomato leaf; a tomato leaf with dry brown bullseye spots.
  \item \textit{tomato leaf with septoria leaf spot}: a tomato leaf with many small round spots with gray centers; tiny brown-edged pale spots scattered over a tomato leaf; a tomato leaf densely speckled with small circular lesions.
  \item \textit{healthy tomato leaf}: a healthy green compound tomato leaf with lobed leaflets; a clean tomato leaf with jagged leaflet edges; a uniformly green tomato leaf.
  \item \textit{tomato leaf with bacterial spot}: a tomato leaf with small dark greasy spots; tiny black lesions with yellow halos on a tomato leaf; a tomato leaf peppered with dark angular spots.
  \item \textit{tomato leaf with late blight}: a tomato leaf with large dark water-soaked blotches; brown-black spreading lesions on a tomato leaf; a tomato leaf with greasy dark patches and pale borders.
  \item \textit{tomato leaf with mosaic virus}: a tomato leaf with a mottled light and dark green mosaic pattern; a curled tomato leaf with patchy yellow-green coloring; a distorted tomato leaf with irregular mottling.
  \item \textit{tomato leaf with yellow leaf curl virus}: a tomato leaf curled upward with yellow edges; small crinkled yellowing tomato leaves; a tomato leaf with yellow margins and a cupped shape.
  \item \textit{tomato leaf with leaf mold}: a tomato leaf with pale yellow patches on top and olive mold below; fuzzy olive-brown mold on the underside of a tomato leaf; a tomato leaf with yellow blotches and velvety mold.
  \item \textit{healthy grape leaf}: a healthy lobed green grape leaf; a broad palmate grape leaf with toothed edges; a clean green vine leaf with five lobes.
  \item \textit{grape leaf with black rot}: a grape leaf with round brown spots with dark borders; tan lesions with black dots on a grape leaf; a grape leaf with circular reddish-brown spots.
\end{itemize}
}
}

{
    \small
    \bibliographystyle{ieeenat_fullname}
    \bibliography{refs}
}